\documentclass[10pt, conference, letterpaper]{IEEEtran}
\IEEEoverridecommandlockouts
\usepackage{cite}
\usepackage{amsmath,amssymb,amsfonts}
\usepackage{algorithmic}
\usepackage{textcomp}
\usepackage{xcolor}
\usepackage{times}
\usepackage{soul}
\usepackage{helvet}
\usepackage{hyperref}
\usepackage{courier}
\usepackage{url}
\usepackage{graphicx}
\usepackage{subfigure}
\usepackage{booktabs}
\usepackage{balance}
\usepackage{multirow}
\usepackage[flushleft]{threeparttable}
\usepackage{soul}
\usepackage{amssymb}% http://ctan.org/pkg/amssymb
\usepackage{pifont}% http://ctan.org/pkg/pifont

\usepackage[ruled,linesnumbered]{algorithm2e}
\SetKw{KwDownTo}{downto}
\SetKw{KwTo}{to}
\SetKw{KwTrue}{true}
\SetKw{KwFalse}{false}
\SetKwInOut{Input}{Input}
\SetKwInOut{Output}{Output}
\SetKw{KwAnd}{and}
\SetKw{KwBreak}{break}

\usepackage{algorithmic}
\usepackage{amsthm}

\newtheorem{theorem}{Theorem}[section]

\newtheorem{proposition}[theorem]{Proposition}
\newtheorem{observation}[theorem]{Observation}

\newcommand{\nguyen}[1]{{\color{black}{#1}}}

\begin{document}
%\bstctlcite{IEEEexample:BSTcontrol} % needed for Only One Author et.al.
%\linenumbers	

\title{CADIS: Handling Cluster-skewed Non-IID Data in Federated Learning with \textbf{C}lustered \textbf{A}ggregation and Knowledge \textbf{DIS}tilled Regularization
}

%, Duc Long Nguyen, , Thanh-Hung Nguyen, Hieu Pham, Truong Thao Nguyen and Phi Le Nguyen

\author{
    \IEEEauthorblockN{Nang Hung Nguyen\IEEEauthorrefmark{1}, 
    Duc Long Nguyen\IEEEauthorrefmark{1},
    Trong Bang Nguyen\IEEEauthorrefmark{1}, 
    Thanh-Hung Nguyen\IEEEauthorrefmark{1},\\
    Huy Hieu Pham\IEEEauthorrefmark{2}, 
    Truong Thao Nguyen\IEEEauthorrefmark{3}\IEEEauthorrefmark{4}\thanks{\IEEEauthorrefmark{4}Corresponding authors}, Phi Le Nguyen\IEEEauthorrefmark{1}\IEEEauthorrefmark{4}}
    \IEEEauthorblockA{\IEEEauthorrefmark{1}School of Information and Communication Technology, Hanoi University of Science and Technology, Hanoi, Vietnam}
    {\{hung.nn184118@sis, long.nd183583@sis, bang.nt190038@sis, hungnt@soict, lenp@soict\}.hust.edu.vn}
    \IEEEauthorblockA{\IEEEauthorrefmark{2}College of Engineering \& Computer Science and VinUni-Illinois Smart Health Center, VinUniversity; hieu.ph@vinuni.edu.vn }
    \IEEEauthorblockA{\IEEEauthorrefmark{2}The National Institute of Advanced Industrial Science and Technology (AIST), Japan; nguyen.truong@aist.go.jp}
}
%\author{\IEEEauthorblockN{1%\textsuperscript{st} Given Name Surname}
%\and
%\and
%\and
%\and
%\and

\maketitle
%%% Code for page numbering
%\thispagestyle{plain}
%\pagestyle{plain}
%%% End  page numbering

\begin{abstract}
Federated learning enables edge devices to train a global model collaboratively without exposing their data.
Despite achieving outstanding advantages in computing efficiency and privacy protection, federated learning faces a significant challenge when dealing with non-IID data, i.e., data generated by clients that are typically not independent and identically distributed.
In this paper, we \nguyen{tackle} %introduce 
a new type of Non-IID data, called cluster-skewed non-IID, discovered in actual data sets. The cluster-skewed non-IID is a phenomenon in which clients can be grouped into clusters with similar data distributions.
By performing an in-depth analysis of the behavior of a classification model's penultimate layer, we introduce a metric that quantifies the similarity between two clients' data distributions without violating their privacy. We then propose an aggregation scheme that guarantees equality between clusters. In addition, we offer a novel local training regularization based on the knowledge-distillation technique that reduces the overfitting problem at clients and dramatically boosts the training scheme's performance. We theoretically prove the superiority of the proposed aggregation over the benchmark FedAvg. Extensive experimental results on both standard public datasets and our in-house real-world dataset demonstrate that the proposed approach improves accuracy by up to 16\% compared to the %state-of-the-art 
FedAvg algorithm.
\end{abstract}


\section{Introduction}
\label{sec:intro}
%\noindent \textbf{Background.}
With the rise in popularity of mobile phones, wearable devices, and autonomous vehicles, the amount of data generated by edge devices is exploding \cite{7498684}. 
%In the future, the amount of data produced is expected to surpass the Internet's capacity \cite{7498684}. 
With the emergence of Deep Learning (DL), edge devices bring endless possibilities for various tasks in modern society, such as traffic congestion prediction and environmental monitoring \cite{8030322, 8270639}.
In the conventional cloud-centric approach, the data from edge devices is gathered and processed at a centralized server \cite{5617062}. This strategy, however, encounters several computational, communication, and storage-related constraints. Critically, the centralization strategy reveals unprecedented challenges in guaranteeing privacy, security, and regulatory compliance \cite{5655240, DOMINGOFERRER201938, MOLLAH201738}.
In such a context, Federated Learning (FL), a novel distributed learning paradigm, emerged as a viable solution, enabling distributed devices (clients) to train DL models cooperatively without disclosing their raw data \cite{fedavg_mcmahan2017communication}. FL prevents user data leakage and decreases server-side computation load.
Each communication round in a standard FL starts with the server transmitting a global model to the clients. Each client then utilizes its own data to train the model locally and uploads the model parameters (or changes), rather than the raw data, to the FL server for aggregation. The server then combines local models to generate an updated version, which is subsequently transmitted to all clients for the next round. This training process terminates once the server receives a desirable model.
\begin{figure}[tb]
    \centering
    \includegraphics[width=0.9\linewidth]{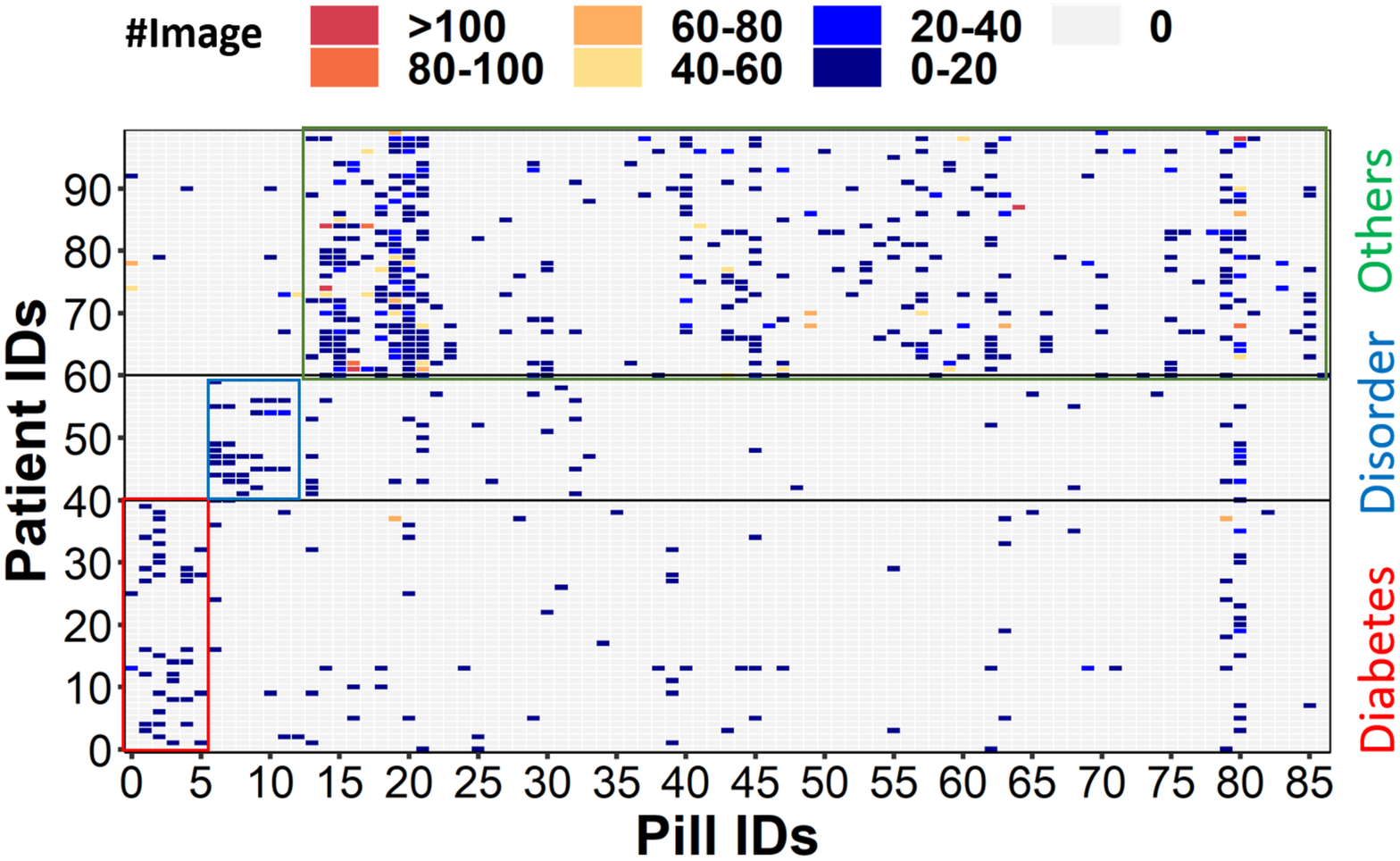}\\
    %\vspace{-10pt}
    \caption{\textbf{Distribution of pill images collected from 100 real patients.} Patients with the same disease usually take similar pills. Data can be classified into three groups: Diabetes (\textcolor{red}{red}), Disorder (\textcolor{blue}{blue)}, and others (\textcolor{green}{green}). \label{fig:pill_distribution}
    %\vspace{-5pt}
    }
\end{figure}
\begin{figure*}[tb]
    %\centering
    %\subfigure[FedAvg]{
    %    \label{fig:fedavg_cfmtx}
    %    \includegraphics[width=0.22\textwidth]{fig/fedavg_cfmtx.eps}
    % }
    %  \subfigure[CADIS]{
    %    \label{fig:CADIS_cfmtx}
    %    \includegraphics[width=0.22\textwidth]{fig/CADIS_cfmtx.eps}
    % }
    %\begin{minipage}{0.4\linewidth}
         \centering
         \small
         \subfigure[Data Distribution]{
            \label{fig:study_distribution}
            \includegraphics[width=0.18\linewidth]{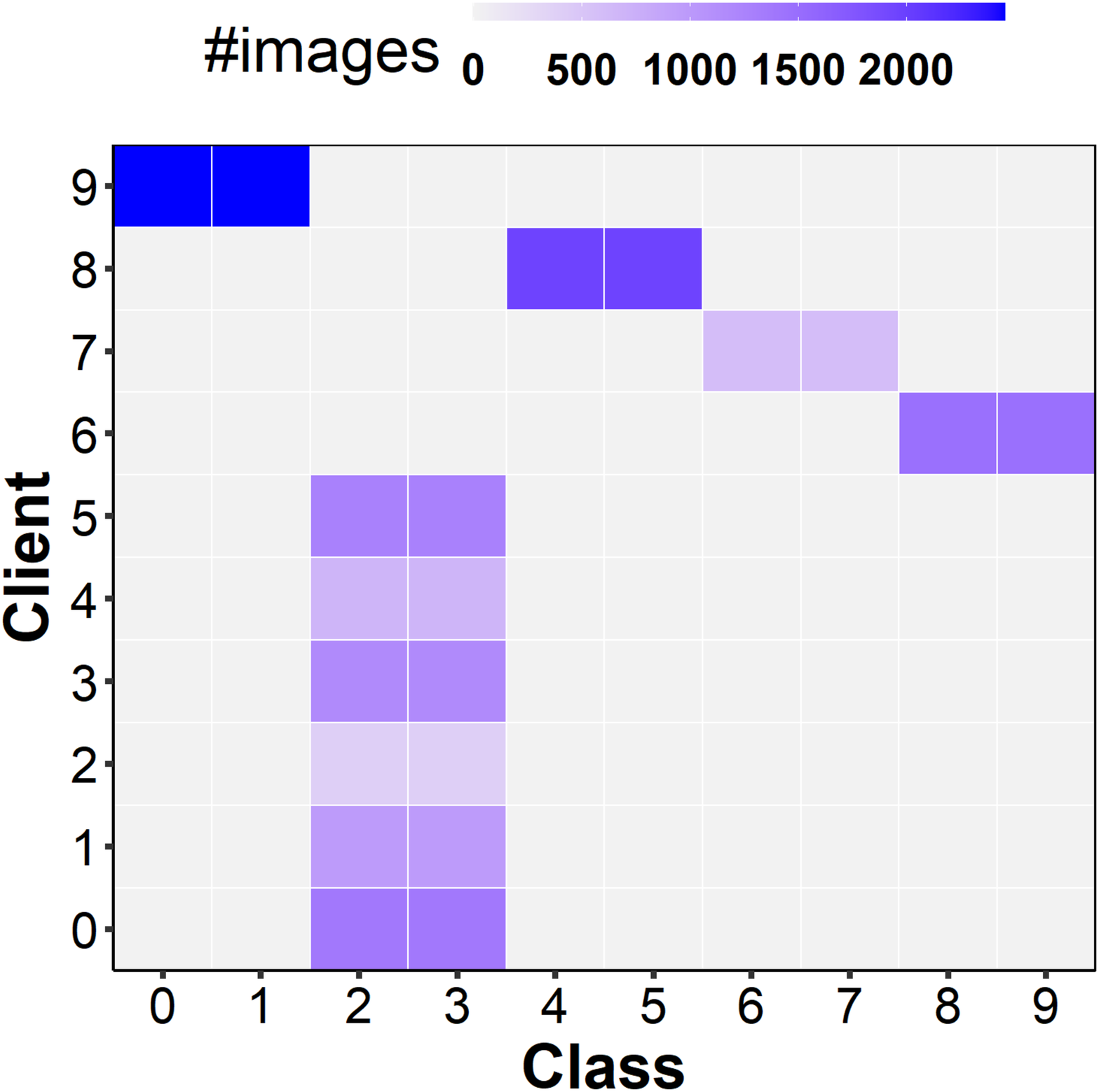}
         }
         \subfigure[FedAvg]{
            \label{fig:study_fedavg}
            \includegraphics[width=0.18\linewidth]{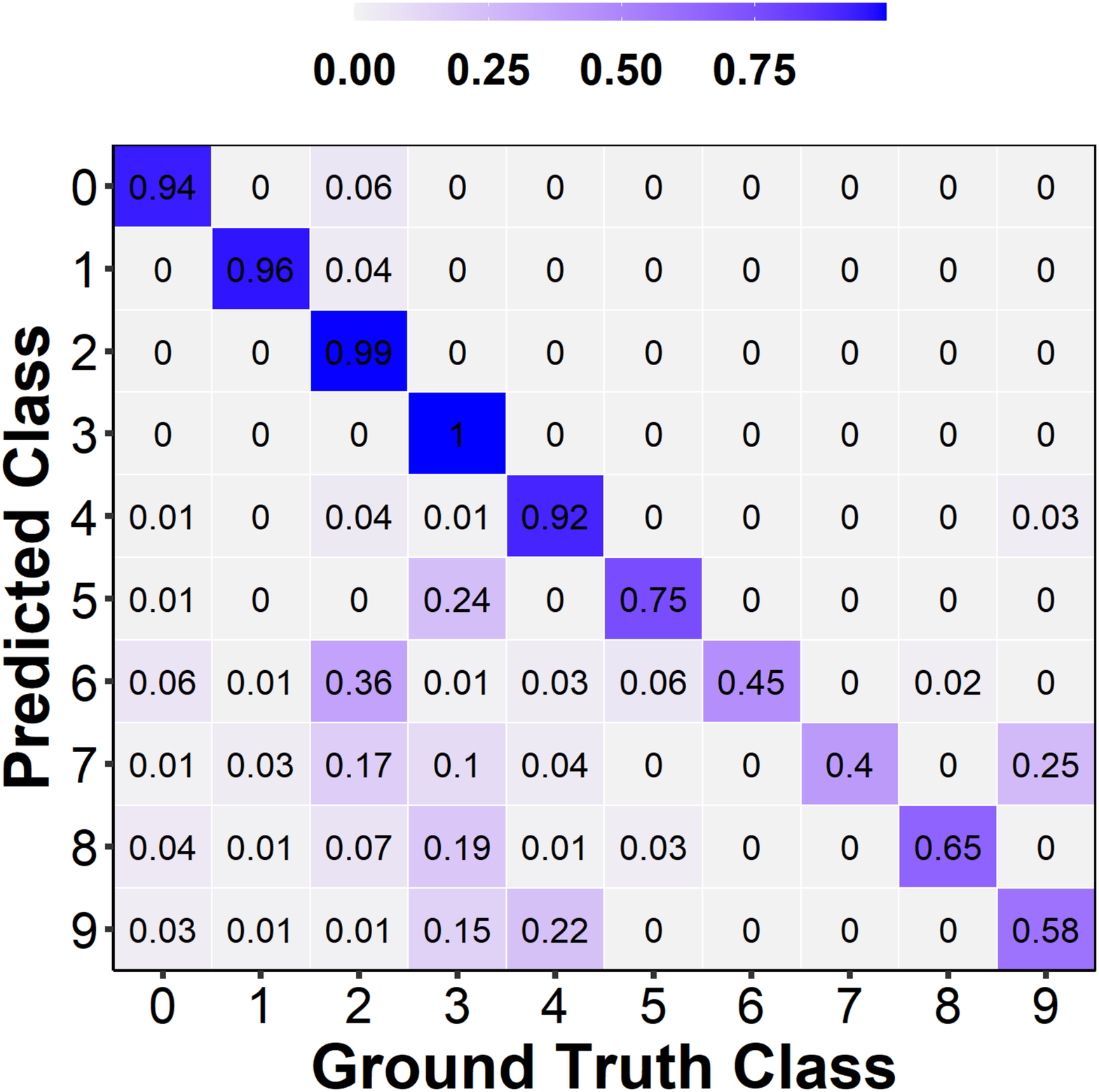}
         }
         \subfigure[FedProx]{
            \label{fig:study_fedprox}
            \includegraphics[width=0.18\linewidth]{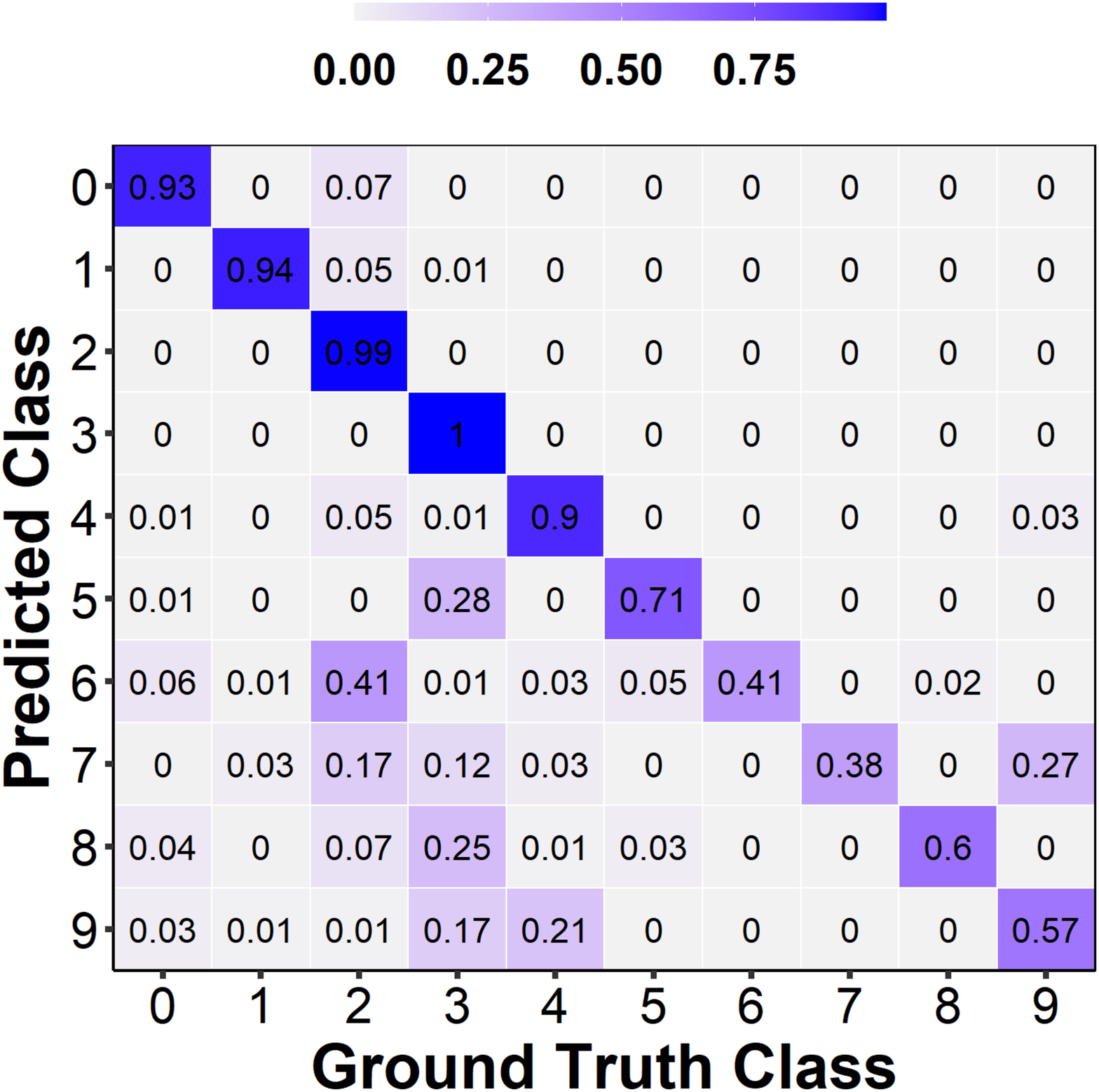}
         }
         \subfigure[FedFA]{
            \label{fig:study_fedfa}
            \includegraphics[width=0.18\linewidth]{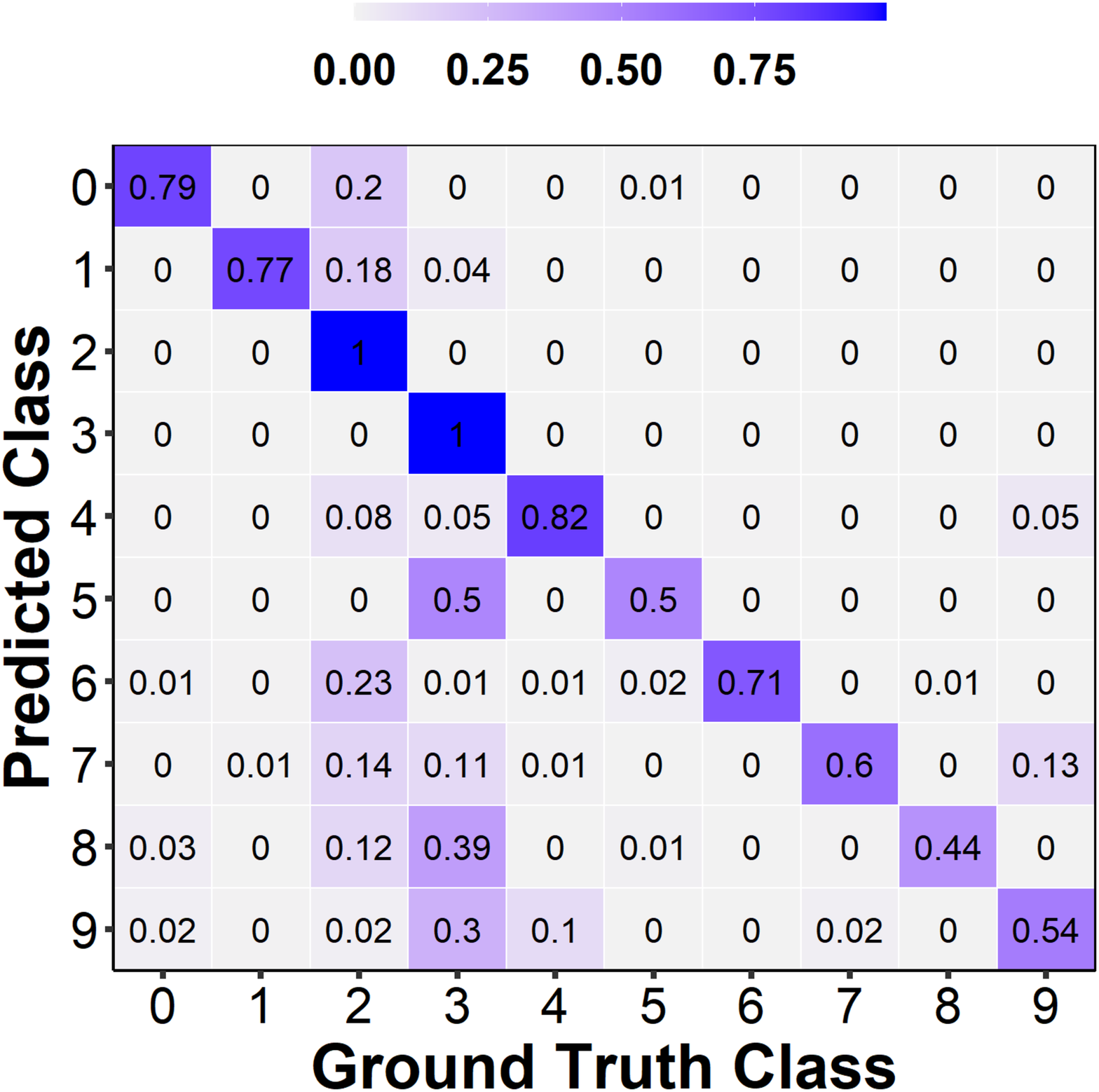}
         }
         \subfigure[CADIS]{
            \label{fig:study_cadis}
            \includegraphics[width=0.18\linewidth]{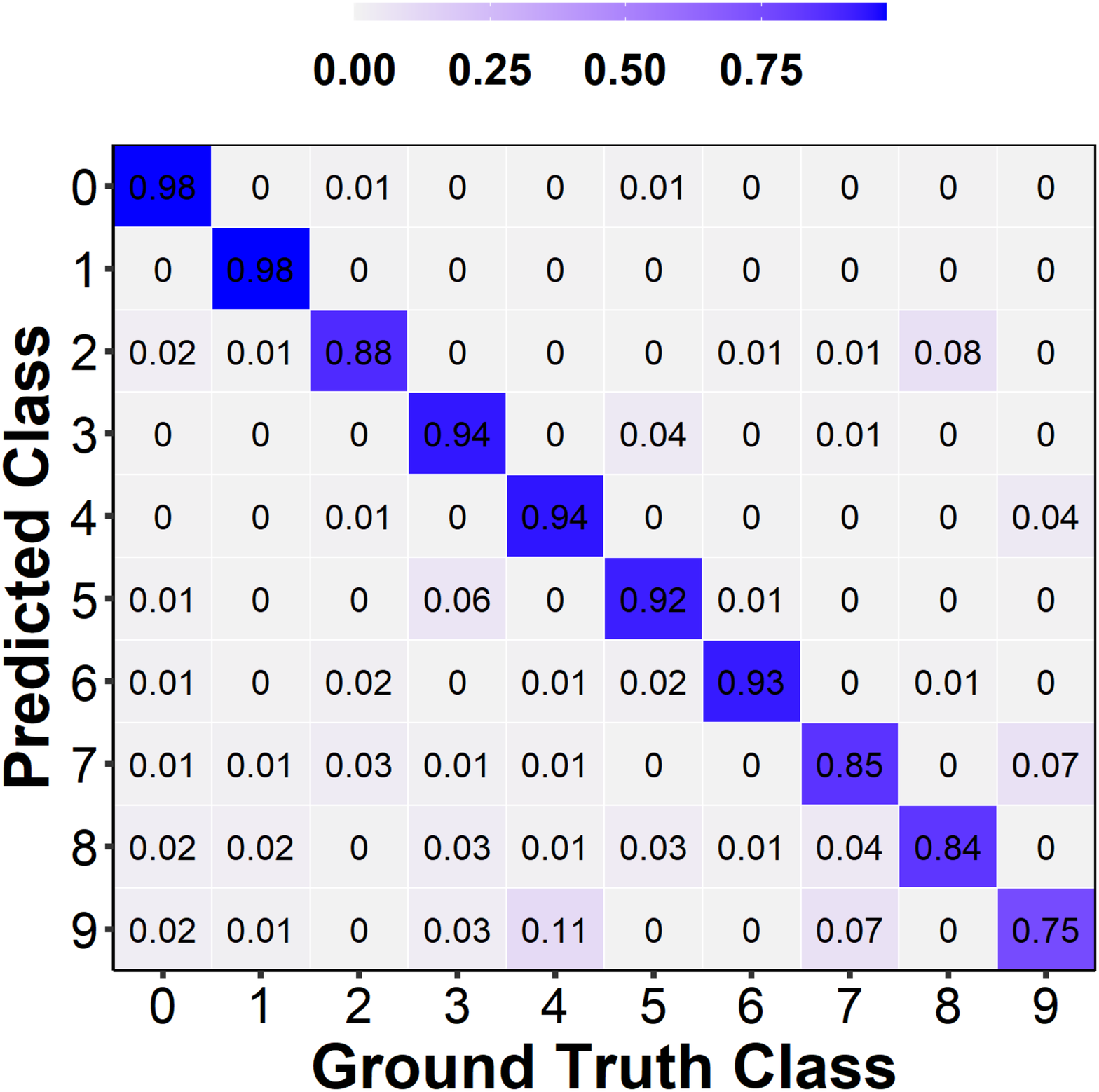}
         }
     % \end{minipage}
     \caption{\nguyen{\textbf{A case-study on the effect of cluster-skew non-IID}. (a): illustration of cluster-skew non-IID on MNIST dataset where client $0 - 5^{th}$ has the same local data distribution that can be grouped into a cluster. (b)-(e) confusion matrix when testing the global model obtained after $100$ training rounds, of FedAvg~\cite{fedavg_mcmahan2017communication}, FedProx~\cite{fedprox_li2020federated}, FedFA~\cite{HUANG2022170}, and CADIS. The value of row $i^{th}$, column $j^{th}$ shows the rate where a samples of class $j$ are predicted as class $i$. Previous works introduce worse performance on the rare classes which belongs to small number of clients, e.g., class $6-9^{th}$. By considering the cardinality of the cluster containing a given client when assign the weight in aggregation at server, CADIS improves the prediction performance in rare classes.
     }}
     \label{fig:cluster_skew_effect}
\end{figure*}
Despite having notable benefits in terms of computing performance and privacy conservation, FL suffers from a significant challenge in dealing with heterogeneous data. In FL, data is generated independently at every device, resulting in highly skewed, non-independent-and-identically-distributed (non-IID) data across clients \cite{MA2022244, zhu2021federated}.
Additionally, the data distribution of each client might not represent the global data distribution.
In 2017, McMahan proposed a pioneer FL model named FedAvg \cite{fedavg_mcmahan2017communication}, which employs SGD for training local models and averaging aggregation at the server-side.
In this work, the authors also mentioned the non-IID issue and argued that FedAvg is compatible with non-IID data.
However, later on, other studies~\cite{li2019convergence,li2020federated}showed that non-IID data substantially impact the performance of FL models (including FedAvg). In particular, non-IID data may slow down the model convergence, destabilize local training at clients, and degrade model accuracy in consequence \cite{9155494, 9496155, 9760086, jiang2022harmofl, zhang2021adaptive}.
%\noindent \textbf{Existing Approaches.} 
Numerous efforts have been devoted to overcoming the non-IID issue, which may be classified into two primary categories: \nguyen{(i) reduce the impact of non-IID data by optimizing the aggregation~\cite{ HUANG2022170, fedfv_wang2021federated, fedadp} or by optimizing the method to select the client each round~\cite{cho2020client,favor_hwangInforcom2020,inforcom2021_fedsens}} on the server-side, and (ii) enhancing training on the client side \cite{fedprox_li2020federated, feddyn_acar2021federated, karimireddy2020scaffold, fednova, ICPP2021_fedCav, BalanceFL2022}.
%\noindent \textbf{Problem Statement and Our Solution.} 
 %despite the fact that many solutions have been proposed,
\nguyen{However,} current research on non-IID faces the two critical issues as follows.

\begin{figure}[tb]
    \centering
    \includegraphics[width=0.85\linewidth]{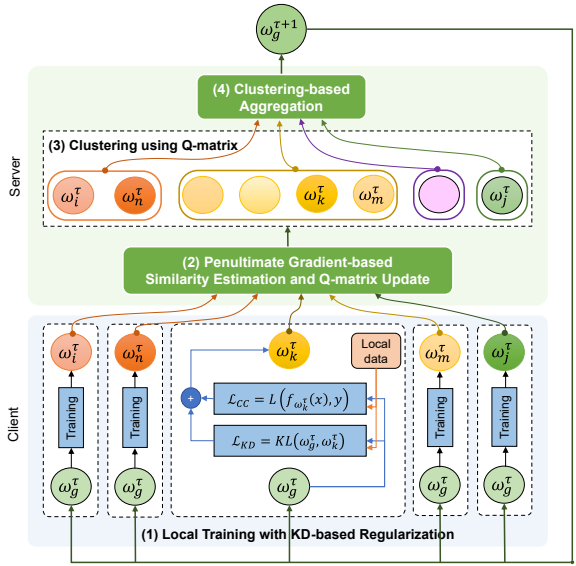}
    \caption{Overview of the proposed CADIS architecture.}
    \label{fig:overview}
    %\vspace{-5pt}
\end{figure}

First, most previous studies have focused only on the \emph{non-identical distribution} aspect, ignoring the \emph{non-independent} feature of the clients' data. 
%\nguyen{Previous studies specifically take into account the label skew non-IID when each client has a fixed number of classes (label size imbalance ~\cite{fedavg_mcmahan2017communication,fedprox_li2020federated,fedat,ICPP2021_fedCav,ijcai2021_Fairness_fedfv,xiao2020averaging}) or when the number of samples of a certain class are distributed to clients using the power-law or Dirichlet distribution (label distribution imbalance~\cite{li2019convergence,IWQOS2021_fedacs,fedfa_huang2020fairness,fednova}).}
In reality, the data collected from clients exhibit a substantial degree of clustering, with many clients having similar labels. For example, consider a pill-recognition FL system in which clients use their taken pill images to train the model \nguyen{(as illustrated in Figure~\ref{fig:pill_distribution})}. Users with the same disease usually have data belonging to identical classes. In other words, clients will be separated into disease-specific categories. In addition, common disease clusters will be considerably larger than other clusters. %Figure~\ref{fig:pill_distribution} illustrates the distribution of pills collected from $100$ real users. 
\nguyen{For example,} the users are classified into three groups \nguyen{in Figure~\ref{fig:pill_distribution}}: diabetic patients, disorder, and others, with the cardinality of the diabetic group being much greater than the disorder group.
%The data demonstrate an imbalance in two manners: 1) Data distribution is disease-specific, i.e., patients with identical diseases often take the same pills, and 2) the number of users belonging to groups is not uniform; for instance, the diabetic group is significantly larger than the other groups, and thus the number of pill images in this group is also dominant. 
To fill in this gap, in this work, besides considering common types of non-IID data as existing studies, we \nguyen{focus on} a new non-IID data that exhibits \emph{non-independent} property of data across clients in this work. Specifically, we \nguyen{tackle} non-IID data having \emph{inter-client correlation}, i.e., clients sharing a common feature can have correlated data. We consider that the global distribution of data labels is not uniform, and data labels frequently are partitioned into clusters. 
The number of clients per group is varied \nguyen{(identified as \textit{cluster-skewed non-IID}~\cite{icml2020_hsieh20a_skewscout, fedDRL})}. 
%We identify this scenario as \textit{cluster-skewed non-IID}. 
%For cluster-skewed data, utilizing the conventional aggregation strategies, which consider the roles of clients equally, will lead to clusters with large numbers of clients dominating the others. 
\nguyen{For cluster-skewed data, utilizing the conventional aggregation strategies, which consider the roles of clients by assigning each client $i$'s local model a weight $p_i$ depend on the intra-client properties
\footnote{\nguyen{ FedAvg~\cite{fedavg_mcmahan2017communication} and its variance methods assign the weighted based on the number of samples in each clients $n_i$, i.e., $p_i = \frac{n_i}{\sum(n_i)}$. Methods that enhances training on client side such as FedProx~\cite{fedprox_li2020federated} give all the clients a same role with $p_i = \frac{1}{\sum(n_i)}$. The approaches optimizing the aggregation adaptively assigned the weights, e.g., based on the client training accuracy and training frequency as in FedFA~\cite{HUANG2022170}.}},  will lead to clusters with large numbers of clients dominating the others. }
To confirm this hypothesis, we have performed a case-study experiment using cluster-skewed data and discovered that in the aggregation process, if we weight clients inversely with the cardinality of the cluster containing them, we can dramatically increase performance compared to vanilla FedAvg \nguyen{(as shown in  Figure~\ref{fig:cluster_skew_effect})}. However, it is crucial to determine how to cluster clients whose dataset is not publicly available. In light of this, we have an important observation that the penultimate layer might provide considerable insights into the training data distribution. Motivated by this fact, we design a novel mechanism to cluster clients based on the data extracted from the penultimate layer of their local models.

Second, the majority of existing approaches  either optimize server-side aggregation~\cite{ HUANG2022170, fedfv_wang2021federated, fedadp} or enhance client-side training efficiency\cite{fedprox_li2020federated, feddyn_acar2021federated}, which results in sub-optimal performance. Therefore, it is crucial to investigate a total solution that simultaneously solves the problem at both the client and server sides. 
We observe that, in reality, the quantity of data possessed by each client is rather small. In addition, due to the non-IID nature, the data distribution of each client does not correspond to the overall data distribution. Therefore, one of the critical dilemmas is that local model trained on the client side quickly over-fits after several epochs~\cite{FL_Trend_2021, pmlr-v97-mohri19a}.
To tackle this issue, we leverage the Knowledge Distillation paradigm and design a regularization term that aims to narrow the gap between the local and global models, thereby preventing the local model from falling into the local minimum.

Figure \ref{fig:overview} depicts the overview of our proposed approach named CADIS (Clustered Aggregation and Knowledge DIStilled Regularization), which consists of four steps: (1) Local training with the aid of KD-based regularization term; (2) Calculating the similarity of clients by utilizing the penultimate layer; (3) Clustering clients into groups; and (4) Aggregating local models using weighted averaging, with the weights determined based on clients' data size and clusters' cardinality. 
%\noindent \textbf{Main Contributions.} 
Our main contributions are as follows.
\begin{enumerate}
    \item We perform a theoretical analysis of the penultimate layer to identify its relationship with the training data. Based on the insights retrieved from the penultimate layer, we offer an approach to quantify the similarity between clients, thereby grouping them into clusters.
    \item We propose a server-side aggregation approach that adequately handles the cluster-skewed non-IID data. The proposed method is applicable to a wide range of non-IID data problems. 
    \item  We provide a knowledge distillation-based regularization term that overcomes the overfitting in the local training process on the client-side.
    \item To demonstrate the superiority of the proposed approach over the state-of-the-art, we conduct comprehensive experiments on the common datasets and our collected real dataset. The results show that our proposal improves the accuracy by up to 16\% compared to the FedAvg.
\end{enumerate}

\section{CADIS - Federated Learning with \textbf{C}lustered \textbf{A}ggregation and
Knowledge \textbf{DIS}tilled Regularization}
\label{sec:CADIS}
%\subsection{Overview of CADIS}
The proposed CADIS framework consists of two main components: Cluster-based aggregation on the server and knowledge distillation-based regularization on the client side. 
Figure \ref{fig:overview} shows the overview of our proposed approach named CADIS. In CADIS, the clients utilize SGD to train the model locally using a loss function composed of the cross-entropy loss and a knowledge distillation-based regularization term. Upon receiving the trained models from the clients, the server leverages information collected from the penultimate layer to assess the similarity between the clients. Specifically, the server maintains a so-called Q-matrix that records the clients' similarities, which are cumulatively updated over communication rounds. Given the similarity of the clients, the server groups them into clusters. Finally, it combines clients' local models using weighted averaging, where each client's weight is determined depending on the quantity of its data and the cardinality of its cluster.

\nguyen{In the following, we first give the details of the aggregation process in Section \ref{subsec:aggregation}. We then present the regularization term in Section \ref{subsec:regularization}. 
Section \ref{sec:evaluation} evaluates the performance of CADIS and compares it to the-state-of-the-art, while the Section~\ref{sec:related_work} presents the related works for dealing with different type of non-IID distributions and different approach of cluster-based federated learning.
Finally, Section \ref{sec:Conclusion} concludes the paper.}

\section{Clustered Aggregation}
\label{subsec:aggregation}
\begin{figure}[t]
    \centering
        \includegraphics[width=\linewidth]{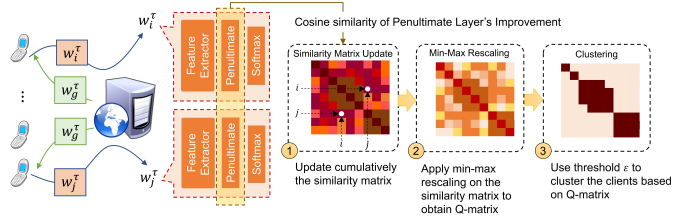}
        \caption{\textbf{An illustration of our proposed clustering algorithm.} In each communication round, the server calculates every client pair's similarity and updates the $Q$-matrix. After that, it partitions the clients into clusters based on their similarities. }
        \label{fig:server_agg}
        %\vspace{-5pt}
\end{figure}
In the following, we first present our proposed cluster-based aggregation formula in Section~\ref{sec_cluster_formula} and then go into the details of our clustering algorithm in Section~\ref{subsec:similarity_calculation}.
\nguyen{Specifically, we introduce an analysis of what the penultimate layer may tell us about the training data distribution in~\ref{subsubsec:Insight}.
Motivated by this finding, we then propose a clustering algorithm based on the improvement of the penultimate layer as shown in Fig.~\ref{fig:server_agg}. 
The main idea is to estimate the clients' data distribution similarity using the improvement of the penultimate layer (~\ref{subsubsec:Similarity_Estimation}) and then partition them according to their similarities (\ref{subsubsec:clustering}). 
Moreover, to speed up the convergence of the similarity matrix, we propose a transitive learning mechanism in Section \ref{subsec:transitive}.
Finally, we provide an analysis on the rate of our clustered aggregation compared to those of the FedAvg in Section~\ref{subsec:convergence}.
}
\subsection{Aggregation Formula}
\label{sec_cluster_formula}
Let $C_1, ..., C_n$ be the $n$ clients.
Suppose that $C_{\tau_1}, ..., C_{\tau_k}$ ($\tau_1, ..., \tau_k \in \{1, ..., n\}$) are the clients participating in the training process at the communication round $t$.
Upon completion of the local training process, these $k$ clients transmit to the server the information of $k$ trained local models, denoted by $\omega^{t}_{\tau_1}, ..., \omega^{t}_{\tau_k}$.
The server will partition $k$ clients into $m_\tau$ clusters using the algorithm provided in Section \ref{subsec:similarity_calculation}.
For each client $\tau_i$, let $M^{t}_{\tau_i}$ be the number of elements of the cluster containing $\tau_i$ at round $t$.
The server then performs weighted aggregation, where client $\tau_i$'s weight, denoted by $\alpha^{t}_{\tau_i}$, is defined as
\begin{equation}
\small
    \label{eq:weight}
    \alpha^{t}_{\tau_i} = \frac{1}{M^{t}_{\tau_i}} \times \frac{n_{\tau_i}}{N},
\end{equation}
where $n_{\tau_i}$ is the number of samples owned by client $\tau_i$, and $N$ is the total samples of all clients. The intuition of this aggregation weight is as follows. 

Clients in the same cluster are supposed to have similar training datasets, resulting in similar locally trained models. Let's consider the following scenario. 
Suppose cluster $A$ has a large number of clients, say fifty, whereas cluster $B$ has a small number of clients, say five. Due to the data similarity, the local training at clients in cluster $A$ produces fifty similar models, and so do the clients in cluster $B$. 
To facilitate the understanding, we refer to $W_A$ and $W_B$ as the ones representing the models of clients in cluster $A$ and cluster $B$, respectively.
If we simply treat all clients equally and aggregate them, then model $W_A$ will have a tenfold greater impact on the global model than model $W_B$.
To equalize the contribution across the clusters, we employ the first term in (\ref{eq:weight}), which is inversely proportional to cluster cardinality.
The second term in (\ref{eq:weight}), 
inherited from FedAvg, is proportional to the number of samples of each client. This term enables clients with more data to contribute more to the global model since clients with more data will, in general, possess more knowledge.
Finally, $\alpha^{t}_{\tau_i}$ is normalized and applied to the client models' weights as follows
\begin{equation}
\small
    \omega_{g}^{t+1} = \sum_{i=1}^{\tau_k}\frac{\alpha^{t}_{\tau_i}}{\sum_{i=1}^{\tau_k}\alpha^{t}_{\tau_i}}\times \omega_{i}^{t}.
\end{equation}
\subsection{Penultimate Layer-assisted Clustering Algorithm}
\label{subsec:similarity_calculation}
%We seek to cluster together clients whose data distributions are similar.
%In the following, we first describe what the penultimate layer may tell us about the training data distribution and then present the details of our clustering algorithm.
%\begin{figure}[tb]
%    \centering
%    \includegraphics[width=1\linewidth]{fig/classification_model2.eps}
%    \caption{\textbf{An example of a neural network for classification task.} 
%    After training with a sample of class $j$, the value of the penultimate layer's $j$-th row increases while those of the remaining rows decrease.}
%    \label{fig:my_model}
%    \vspace{-5pt}
%\end{figure}
\subsubsection{Insights of the Penultimate Layer}
\label{subsubsec:Insight}
Let us consider a typical deep neural network $\mathcal{M}$ for a classification task, consisting of a feature extractor and a classifier that is trained using the cross entropy loss by SGD method. 
We assume that the classifier comprises of a dense layer, represented by $W$, followed by a softmax layer. %(Fig.~\ref{fig:my_model}). 
Here, we give the mathematical supports for the non-bias case because the maths can be easily extended by append a constant $1$ to the sample vector $x$.
Suppose there are $v$ classes, denoted by $1, ..., v$.
We have the following observations.
% standard classification task in which the neural network is composed of two successive components: the feature extractor (encoder) and the classifier (decoder), as depicted in Figure~\ref{fig:my_model}. For simplicity, the classifier consists of a dense layer only whereas the feature extractor can be designed using a variety of architectures.
% We then sample an input $(x, y)$ where $x$ is the data and $y$ is the corresponding groundtruth label, in this specific case, $y = j$. For convenience, we encode $y$ as a one-hot vector whose $j$ element $y_j$ is marked $1$, elsewhere $0$.
\begin{proposition}
\label{prop:pen_shift}
Suppose $W=[\textbf{w}_1, ..., \textbf{w}_v]$, where $\textbf{w}_i$ is the $i$-th row of $W$. Let ${x}$ be a sample with the groundtruth label of $j$, $j\in \{1, ..., v\}$, and $y\in \mathbb{R}^v$ be the one-hot vector representing $j$. 
After training the model $\mathcal{M}$ with sample $(x,y)$, the values of all items in $\textbf{w}_j$ increase while that of the other rows decrease. 
\end{proposition}
%We leave the proof in the supplementary due to the space constraint.

\begin{figure}[tb]
    % \centering
    % \begin{minipage}{0.6\linewidth} %0.42
        \centering
     	\subfigure[Training with MNIST]{
     	 \centering
        \includegraphics[width=0.45\linewidth]{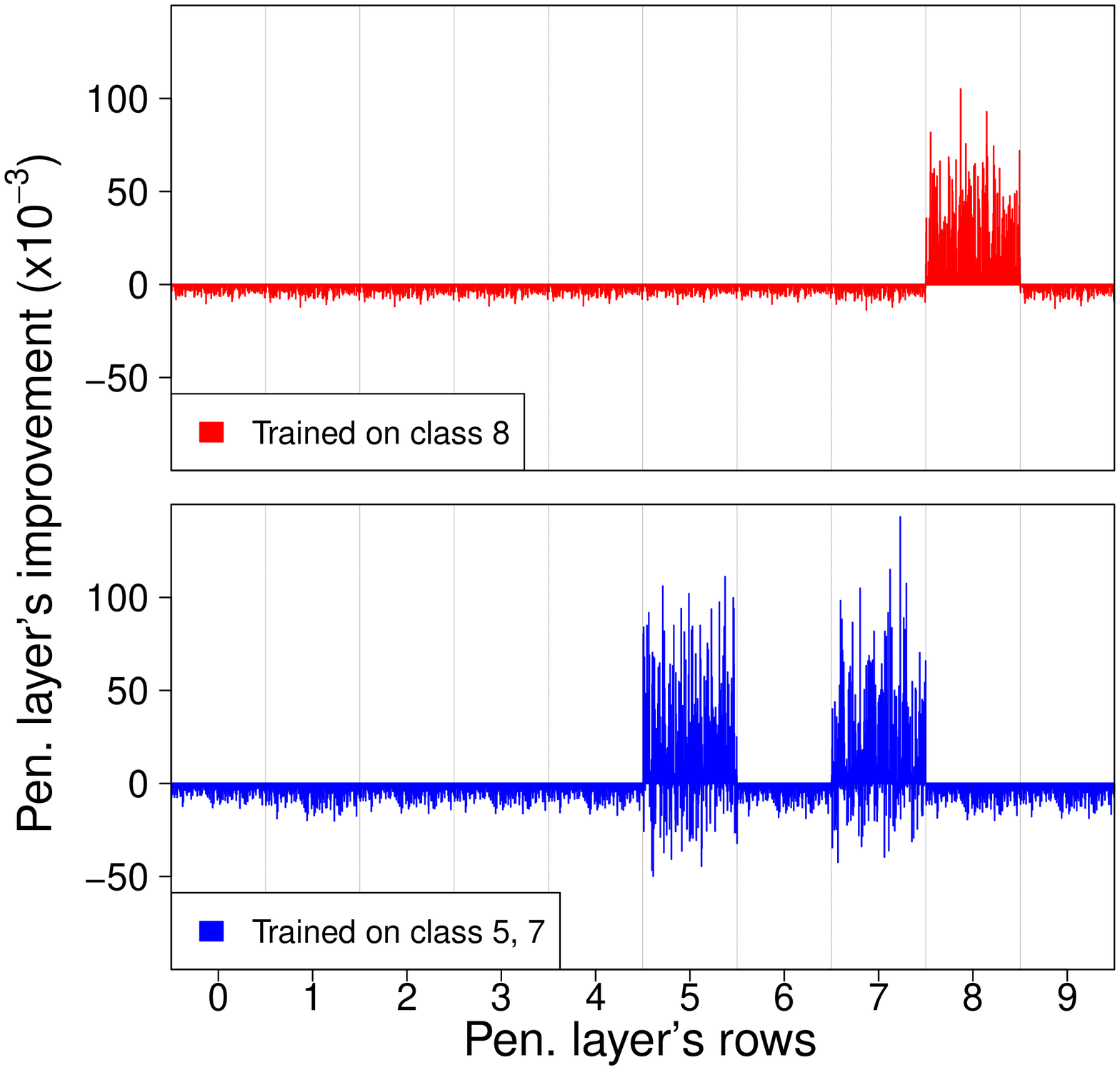}
        \label{fig:mix_class}
        }
     	\subfigure[Training with Pill dataset]{
     	 \centering
        \includegraphics[width=0.42\linewidth]{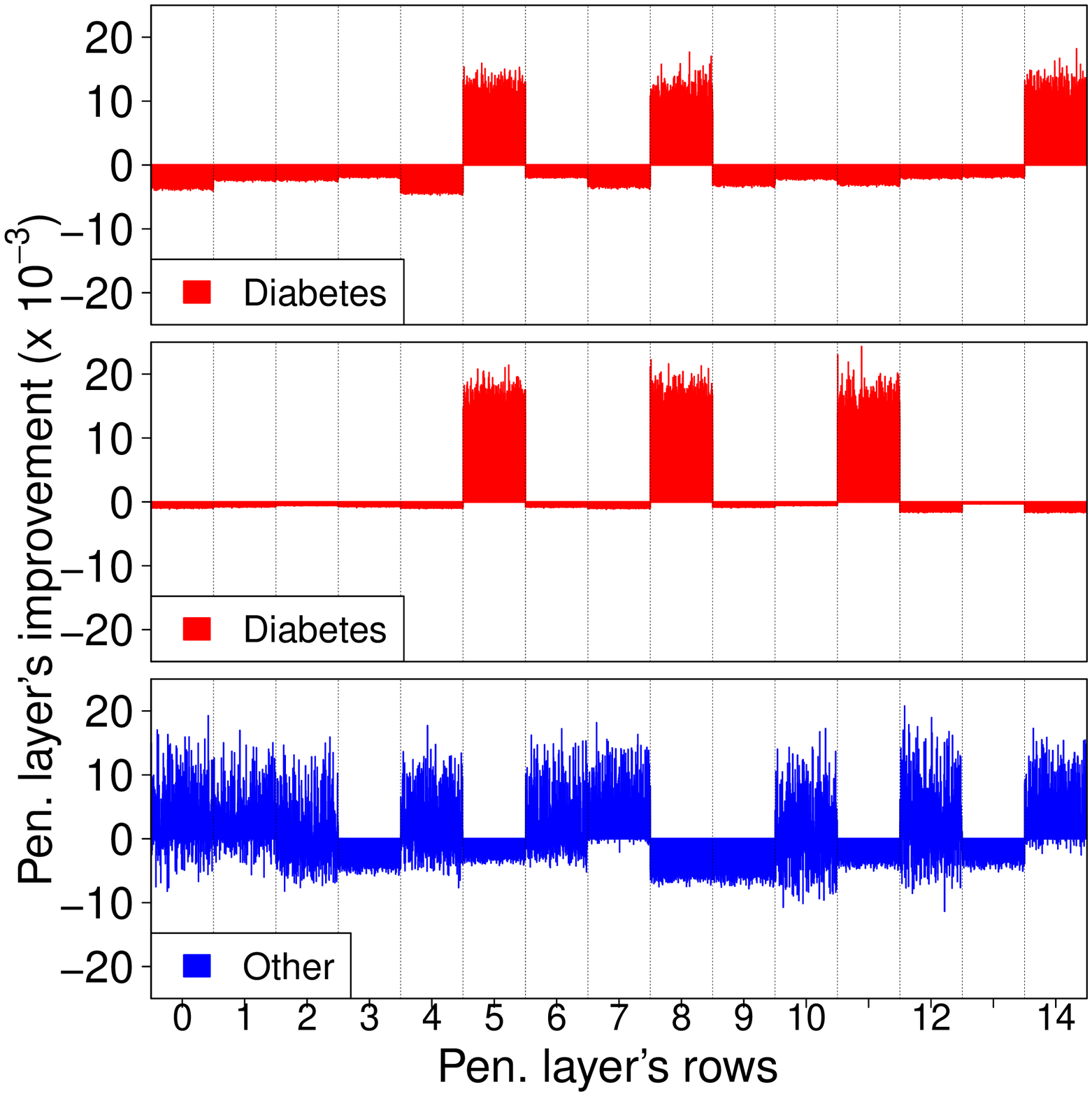}
        \label{fig:multi_class} 
     	 }
     	 %\vspace{-0.2cm}
     	 \caption{\textbf{The behavior of the penultimate layer.} The rows corresponding to the untrained classes decrease. \label{penultimate_layer_gradient}
       %\vspace{-5pt}
     	 }
    % \end{minipage}
    % \hfill
    % \begin{minipage}{0.8\linewidth} %0.55
    %     \centering
    %     \includegraphics[width=1\linewidth]{fig/server_aggregation.eps}
    %     \caption{\textbf{An illustration of our proposed clustering algorithm.} In each communication round, the server calculates every client pair's similarity and updates the $Q$-matrix. After that, it partitions the clients into clusters based on their similarities. }
    %     \label{fig:server_agg}
    % \end{minipage}
    %   \vspace{-10pt}
\end{figure}

Figure~\ref{fig:mix_class} illustrates an intuition for Proposition \ref{prop:pen_shift}. In the upper sub-figure, we trained the model with a sample belonging to class 8 and measured the change of the penultimate layer. 
It can be observed that the 8-th exhibits positive growth, whereas the remaining rows have negative values. 

%\begin{proof}
\noindent \textbf{Sketch Proof.}
Let $R = R(x) \in \mathbb{R}^u$ denote the representation of $x$. For the sake of the arguments, we assume that all items $R_i$ of $R$ are non-negative (attainable with the most popular Sigmoid or ReLU activation functions). Let us denote by $L(x) \in \mathbb{R}^v$  the logits of $x$, then $L(x)$ is defined as follows
\begin{equation}
\small
    L(x) = W \cdot R = \begin{bmatrix}
       R_1 w_{11} + R_2 w_{12} + \dotsc + R_u w_{1u}           \\[0.3em]
       R_1 w_{21} + R_2 w_{22} + \dotsc + R_u w_{2u}           \\[0.3em]
       \vdots\\[0.3em]
       R_1 w_{v1} + R_2 w_{v2} + \dotsc + R_u w_{vu}
     \end{bmatrix}.
\end{equation}
Let $p(x)$ be the prediction result which is the output of the softmax layer, then the probability of sample $x$ being classified into class $j$, i.e., $p_j(x)$, is determined by the following formula
\begin{equation}
\small
    % p(y=j|x) = p(y_j = 1)
    p_j(x) = \frac{e^{L_j}}{\sum_{i=1}^{v} e^{L_i}}.
    %\vspace{-5pt}
\end{equation}
The cross entropy loss concerning sample $(x,y)$ is given by
\begin{equation}
\small
    \mathcal{L}(p(x), y) = \sum_{i=0}^{v} {y_i \log \bigg(\frac{1}{p_i(x)}\bigg)}.
    %= \sum_{i=0}^{v} {y_i \log \bigg(\frac{1}{P_i}\bigg)}.
\end{equation}
Let $w_{rc}$ be the item at row $r$ and column $c$ of $W$, then the gradient of the loss $\mathcal{L}(p(x), y)$ with respect to $w_{rc} \in W$ is %defined as
\begin{equation}
\small
    \frac{\partial \mathcal{L}}{\partial w_{rc}} = \sum_{i=1}^{v}{\bigg(\frac{\partial \mathcal{L}}{\partial p_i(x)} \cdot 
    \bigg(\sum_{k=1}^{v}\frac{\partial p_i(x)}{\partial L_k} \cdot \frac{\partial L_k}{\partial w_{rc}} \bigg)\bigg)}. 
    \label{eqn:grad}
\end{equation}
We have
\begin{equation}
\small
    \frac{\partial \mathcal{L}}{\partial p_i(x)} = -\frac{1}{\ln{10}}\frac{y_i}{p_i(x)} = 
    \begin{cases}
        -\frac{1}{\ln{10}}\frac{y_j}{p_j(x)} & \text{if $i=j$}\\
        0 & \text{if $i\neq j$}
    \end{cases},
    \label{eqn:grad1}
\end{equation}

\begin{align}
\small
    \label{eqn:grad2} \frac{\partial p_j(x)}{\partial L_k} &= %pi(x) -> pj(x)
    \begin{cases}\small
        p_j(x)(1-p_k(x)) & \text{if $k=j$} \\
        -p_j(x) p_k(x) & \text{otherwise}
    \end{cases};\\
    \frac{\partial L_k}{\partial w_{rc}} & = 
    \begin{cases}\small
        R_c & \text{if $k=r$} \\
        0 & \text{otherwise}
    \end{cases}. %R_c \text{  }\forall r,c,k.
    \label{eqn:grad3}
\end{align}
% and 
% \begin{equation}
% \end{equation}
\noindent From (\ref{eqn:grad1}, \ref{eqn:grad2}) and (\ref{eqn:grad3}), we deduce that 
% log10 -> ln10
\begin{equation}
\small
    \frac{\partial \mathcal{L}}{\partial w_{rc}} = 
    \begin{cases}
        \frac{-1}{\ln{10}}y_j (1 - p_j(x)) R_c & \text{if $r = j$},\\
        \frac{1}{\ln{10}}y_j p_r(x) R_c & \text{otherwise.}
    \end{cases}
\end{equation}
As $y_j = 1$, $p_i(x) > 0$ and $R_c > 0$  ($\forall i, c$), when applying the gradient descent, the values of the $j$-th row of $W$ increase while those on all other rows decrease.

Proposition \ref{prop:pen_shift} can be generalized (with slightly more work)
to the case where multiple labels being trained during the training process. Figure~\ref{fig:mix_class} depicts an illustration for the general case. In the lower sub-figure, we trained the model with samples belonging to classes $5$ and $7$. As seen, only the rows $5, 7$ may contain positive values, while values of the remaining rows are strictly negative. From this proposition we come up to the following observation.

% % This insight from \textit{proposition \ref{prop:pen_shift}} helps uncovering some key aspects about the data set on which the model was trained on, especially the labels of the trained data set. Let $\mathcal{D} := \{(n_i, g_i)\}|i \in [1, m]$ denote the data set with $m$ labels where $n_i$ is the number of samples of class $g_i$. 
% \begin{proposition}
% Given data set $d \subset \mathcal{D}$, $d := \{(n_i, g_i)\}$, and its label set $G := \{g_i\}$ where $|d| < m$. And the penultimate layer of the model before being trained $(W)$ and after being trained $(W')$ on $d$. With $w_{kc} \in W$ and $w'_{kc} \in W'$ then $\Delta w_{kc} = w'_{kc} - w_{kc} < 0$ $\forall k \notin G$.
% \label{prop:general}
% \end{proposition}
% \begin{proof}
% See Appendix~\ref{appendix:pen_2}
% \end{proof}
% The intuition from \textit{proposition \ref{prop:general}} is that \textbf{\textit{when we observe the cumulative gradients of each row of the penultimate layer's parameter matrix, if on row $i$ exists any positive value then the data set on which the model was trained must contains samples of label $i$}}.
\begin{observation}
\label{observation}
By analyzing the improvement of the penultimate layer, we may identify whether the training data comprises samples from a particular class.
Specifically, the training data consists of class $j$'s samples if and only if the improvement of the $j$-th row of the penultimate layer's matrix is not negative (i.e., at least one item in the $j$-th row gets higher after training).
\end{observation}

Figure~\ref{fig:multi_class} depicts our observation \ref{observation} in the context of the real-world scenario. Specifically, we train three local models using three pill datasets, two of which contain images of pills taken by diabetic patients and the other by a normal user. The figure demonstrates that the improvement of the penultimate layers of the two diabetic patients is comparable, whereas that of the normal user is clearly different.

%Motivated by this finding, we propose a clustering algorithm based on the improvement of the penultimate layer as shown in Fig.~\ref{fig:server_agg}. 
%The main idea is to estimate the clients' data distribution similarity using the improvement of the penultimate layer and then partition them according to their similarities. 
\subsubsection{Similarity Estimation}
\label{subsubsec:Similarity_Estimation}
Let $\mathcal{M}_i$, and $\mathcal{M}_j$ be two models locally trained by client $C_i$ and $C_j$ using their respective datasets $D_i$ and $D_j$. 
We seek to estimate the similarities of the distributions of $D_i$ and $D_j$ by using the information obtained from the penultimate layers of $\mathcal{M}_i$ and $\mathcal{M}_j$.
To ease the presentation, in the following, we use the term \emph{similarity of client $C_i$ and $C_j$} to indicate the similarity between $C_i$ and $C_j$'s data distributions.
We encounter the following two significant challenges.
First, in the FL training methodology, only a portion of clients engage in the training process during each communication round. Consequently, it is impossible to gather information on the penultimate layers of all clients concurrently.
Second, we observe that the change in the penultimate layer throughout each communication round is negligible. It is thus impossible to determine similarity using the raw improvement of the penultimate layer.

To address the first issue, the server will maintain a so-called similarity matrix whose each item $s_{ij}$ depicts the estimated similarity between client $C_i$ and $C_j$. 
In each communication round $t$, for each pair of clients $(C_i, C_j)$ participating in that round, the server estimates the instance similarity $s^{t}_{ij}$ of $C_i$ and $C_j$, which depicts the similarity of the training data of $C_i$ and $C_j$ at round $t$. $s^{t}_{ij}$ is defined by the following formula
\begin{equation}
\small
    s^t_{ij} = \frac{(W^{t}_i -W_g^{t})^T \cdot (W^{t}_j - W_g^{t})}{\|W^{t}_i - W_g^{t} \|\|W^{t}_j-W_g^{t}\|},
    \label{eqn:sim}
\end{equation}
where $W^{t}_i$ and $W^{t}_j$ are the penultimate layers of $\mathcal{M}_i$ and $\mathcal{M}_j$ at round $t$, while $W_g^{t}$ is the penultimate layer of the global model that the server delivered to the clients at the beginning of round $t$. Note that, $W^{t}_i -W_g^{t}$ and $W^{t}_j - W_g^{t}$ are the improvements of $C_i$ and $C_j$'s local models' penultimate layers after training at round $t$, respectively. 
Therefore, $s^t_{ij}$ indicates the cosine similarity between the penultimate layers' improvements.

As the instance similarity $s^t_{ij}$ may not accurately reflect the actual similarity between clients, we utilize $s^{t}_{ij}$ to update the cumulative similarity $s_{ij}$ in the similarity matrix to achieve accurate estimates.
Specifically, $s_{ij}$ is updated as
\begin{equation}
\small
    s_{ij} \leftarrow \frac{f^t_{ij}}{f^t_{ij}+1} s_{ij} + \frac{1}{f^t_{ij}+1} s^{t}_{ij},
\end{equation}
where $f^t_{ij}$ represents the total times $C_i$ and $C_j$ have participated in the same communication round up to round $t$.
%Moreover, to speedup the convergence of the similarity matrix, we propose a transitive learning mechanism (Section \ref{subsec:transitive}).

The second issue, namely the incremental improvement of the penultimate layer, results in the similarity value of all client pairs rapidly converging to $1$.
To this end, our solution is to use the min-max rescaling on the similarity matrix to obtain a so-called $Q$-matrix. \\[-0.3cm]
% whose every term $q_{ij}$ is derived from the similarity matrix as follows.
% \begin{equation}
% \small
%     q_{ij} \leftarrow \frac{s_{ij}-s_{min}}{s_{max}-s_{min}},
% \end{equation}
% where $s_{min}$ and $s_{max}$ are the minimum and maximum of items in the similarity matrix. \\

\subsubsection{Client Clustering}
\label{subsubsec:clustering}
Given the $Q$-matrix at a communication round $t$, the server uses a binary indicator $u_{ij}$ to determine whether clients $C_i$ and $C_j$ belong to the same cluster as in Equation~\ref{eq:eq13}, where $\varepsilon$ is updated upward after every communication round.
Note that as $q_{ij}$ is adjusted every round, $u_{ij}$ is also updated over communication rounds, but it will converge after some rounds
\begin{equation}
\small
    \label{eq:eq13}
    % \small
    u_{ij} = 
    \begin{cases}
        1, &\text{if } q_{ij} \geq \varepsilon; \\
        0, &\text{otherwise}.
    \end{cases}
\end{equation}

%The theoretical study of the convergence rate of the $Q$-matrix is provided in the Supplementary \cite{supplementary}.
\subsection{Enhancing the Similarity Matrix with Transitive Learning}
\label{subsec:transitive}
In the FL training methodology, in each round, there is only a portion of clients participating in the training process. Therefore, it requires significant time for the similarity matrix to converge. 
To speedup the convergence, we propose an algorithm to estimate the similarity of two arbitrary clients $C_i$ and $C_j$ via their similarities with other clients. 
We notice that cosine similarity possess a transitive characteristic which is reflected by the following theorem~\cite{schubert2021triangle}.
\\[-0.5cm]
\begin{theorem}
    \label{eq:cosine_transitive}
    Let $s_{x,y}$ denote the cosine similarity of two vectors $x$ and $y$. Given three arbitrary vectors $x, y$ and $z$, then their cosine similarities satisfy the following inequality
    \begin{equation}
    \begin{split}
    \small
    \nonumber
         s_{a,b}s_{b,c} - \sqrt{\left(1-s_{a,b}^{2}\right)\left(1-s_{b,c}^{2}\right)} \leq s_{a,c} \\
         \leq s_{a,b}s_{b,c} + \sqrt{\left(1-s_{a,b}^{2}\right)\left(1-s_{b,c}^{2}\right)}.
    \end{split}
    \end{equation}
\end{theorem}
Motivated by this theorem, we utilize the Gaussian distribution with the mean of $s_{ip}s_{jp}$ and deviation of $\frac{\sqrt{(1-s^{2}_{ip})(1-s^{2}_{jp})}}{3}$, denoted as $\mathcal{N}_{(s_{ip},s_{jp})}$, to estimate the value of $s_{ij}$. 
Accordingly, for every client pair $(C_i, C_j)$ which does not co-occurence in a communication round $t$, the server will find all clients $C_p$ such that the deviation of $\mathcal{N}_{(s_{ip},s_{jp})}$ is less than a threshold $\gamma$, i.e.,  $\frac{\sqrt{(1-s^{2}_{ip})(1-s^{2}_{jp})}}{3} < \gamma$ ($^*$). 
For each such a client $C_p$, we denote by $s_{ij,p}$ a random number following the distribution $\mathcal{N}_{(s_{ip},s_{jp})}$.
The final estimated value for $s^t_{ij}$ is the average of $s_{ij,p}$ for all $p$ satisfying condition ($^*$).
\begin{theorem}
\label{pro:similarity}
Let $n$ be the total number of clients and $k$ be the number of clients participating in a communication round. Let $\delta$ be the expected number of communication rounds needed to estimate the similarity of all client pairs.
Then, $\delta \leq 1 + \sum_{i=k}^{n-1}\frac{\binom{n}{k}}{\binom{n}{k}-\binom{i}{k}}$.
\end{theorem}
\noindent \textbf{Sketch Proof.} 
%Due to the space constraints, we only provide a sketch proof. 
We denote $S_i$ $(\forall i \in [0, n])$ as a random variable representing the number of communication rounds needed for all clients participate in the training at least one time, given $i$ clients have already participated in training so far.
Then, $\delta$ equals the expected value of $S_0$, i.e., $E(S_0)$.
$E(S_0)$ can be determined recursively as follows
\begin{equation}
\label{eqn:S0}
\begin{small}
    \begin{cases}
        & E(S_0)  = 1 + E(S_k) , \\
        %& {E}(S_1)  = a_{10} (1 + {E}(S_k) ) + a_{11} (1 + {E}(S_{k+1}) ), \\
        %&\dots\\
        & {E}(S_i)  = \sum_{j=0}^i a_{ij} (1 + {E}(S_{k+j})), \forall i\\
       % &\dots\\
        & {E}(S_n)  = 0,
    \end{cases}
\end{small}
\end{equation}
where $a_{ij}$ is the transitioning probability from state $S_i$ to $S_{k+j}$ defined by $a_{ij} = \frac{\binom{n-i}{k+j-i} \times  \binom{i}{i-j}}{\binom{n}{k}}$.
We have $\sum_{j=0}^{i}a_{ij} = 1$, and $E(S_i) \geq E(S_{i+1})$ ($\forall i$).
Moreover, when $i \geq k$, we have $a_{ij} = 0$ ($\forall j < i-k$).
Therefore, $\sum_{j=0}^{i}a_{ij} = \sum_{j=i-k}^{i}a_{ij} = 1$. 
Accordingly, 
\begin{small}
\begin{align*}
    E(S_i) &= 1 + \sum_{j=i-k+1}^{i} {a_{ij} \times E(S_{k+j}) + a_{i(i-k)} \times E(S_i)} \\
    & \leq 1 + \sum_{j=i-k+1}^{i} {a_{ij} \times E(S_{i+1}) + a_{i(i-k)} \times E(S_i)} \\
    & = 1 + (1-a_{i(i-k)}) \times E(S_{i+1}) + a_{i(i-k)} \times E(S_i).
\end{align*}
\end{small}
It can be deduced that 
\begin{small}
\begin{align*}
    E(S_i) & \leq E(S_{i+1}) + \frac{1}{1-a_{i(i-k)}} = E(S_{i+1}) + \frac{\binom{n}{k}}{\binom{n}{k} - \binom{i}{k}} \\
    \Rightarrow E(S_k) & \leq E(S_n) + \sum_{i=k}^{n-1}\frac{\binom{n}{k}}{\binom{n}{k} - \binom{i}{k}} = 0 + \sum_{i=k}^{n-1}\frac{\binom{n}{k}}{\binom{n}{k} - \binom{i}{k}} \\
    \Rightarrow E(S_0) & = 1 + E(S_k) \leq 1 + \sum_{i=k}^{n-1}\frac{\binom{n}{k}}{\binom{n}{k} - \binom{i}{k}}.
\end{align*}
\end{small}
% \begin{equation}
%     a_{ij} = \frac{\binom{n-i}{k+j-i} \times  \binom{i}{i-j}}{\binom{n}{k}}.
%     \label{eqn:a}
% \end{equation}
%Due to the space constraints, we leave the proof in the Supplementary \cite{supplementary}.

%\input{algo/clustering}

% $Q$-matrix updating scheme presented above works excellently in the case where all clients participate in a round. However, for other cases where only a proportion of the clients participate every round, $Q$-matrix learning process is slightly longer since the similarity between client $c_i$ and $c_j$ $(\forall j \neq i)$ can only be computed when the two clients appear at a same round. Instead, we use the transitive property of cosine similarity proposed by \cite{schubert2021triangle} to speed up the filling of $Q$-matrix. In details, let $A = s(a, b)$ denote the similarity between $a$ and $b$, similarly $B = s(b, c)$ and $C = s(a, c)$, then we have the the triangle inequality:
% \begin{equation}
%     AB - \sqrt{(1-A^2)(1-B^2)} \leq C \leq AB + \sqrt{(1-A^2)(1-B^2)}.
% \end{equation}
% From which, we can estimate using Gaussian approximation:
% \begin{equation}
%     C \approx \mathcal{N}\bigg(AB, \frac{\sqrt{(1-A^2)(1-B^2)}}{3}\bigg).
%     \label{eqn:transition}
% \end{equation}
% By applying Equation~\ref{eqn:transition}, the $Q$-matrix is assured filled when all clients participate at least once, which substantially reduces the number of communication rounds required. We formulate a recursive formula to compute the expected number of rounds for $Q$-matrix to be filled in various settings, see Appendix~\ref{appendix:Q_convg}.

\subsection{Convergence Analysis}
\label{subsec:convergence}
Finally, we have the following finding on the convergence rate of \nguyen{proposed clustered aggregation} compared to FedAvg. 
\begin{proposition}
\label{pro:loss}
Once converged, the inference loss of the global model achieved by CADIS's aggregation process is smaller than those generated by FedAvg
\begin{equation}
\small
    \mathcal{L}_{FedAvg} - \mathcal{L}_{CADIS}
    \geq 0,
\end{equation}
\end{proposition} 
\noindent where $\mathcal{L}_{FedAvg}$ and $\mathcal{L}_{CADIS}$ indicate, respectively, the losses of the converged global models derived by FedAvg and CADIS. Due to space constraints, in the following, we provide a sketch proof when there are clusters among the clients. 

%\begin{proof}
\noindent \textbf{Sketch Proof.}
To simplify, we consider a FL with three clients $C_1, C_2, C_3$, in which $C_1$ and $C_2$ belong to a cluster and $C_3$ does not.
As loss functions are usually convex ones and have at least one minimum, we consider the simple loss functions for $C_1, C_2, C_3$ as $f_i = a_i z^2 +b_iz, (a_i > 0, i= 1, 2, 3)$.
% \begin{equation}
% \small
%         f_i = a_i z^2 +b_iz, \text{  } a_i > 0, ~~ i= 1, 2, 3.
% \end{equation}
Let $D_i$ be the dataset owned by $C_i$. As $C_1$ and $C_2$ belong to the same cluster, we assume that $C_1$ and $C_2$ are similar. 
Let $\mathcal{D}$ be the dataset whose distribution is identical with our targeted data, then $\mathcal{D}$ can be form by taking a half of $D_1$, $D_2$ and all of $D_3$. 
Accordingly, we can prove that, the optimal loss function when we train a global model with the single set of $\mathcal{D}$ is given by 
%$f^* = \frac{1}{4}(f_1 + f_2) + \frac{1}{2} f_3$.
\vspace{-5pt}
\begin{equation}
    \label{eq:global_optimal}
\small
    f^* = \frac{1}{4}(f_1 + f_2) + \frac{1}{2} f_3.
\end{equation}
Let  $z_{i}^{t,E}$ ($E$ is the number of training epochs) be the trained model that $C_i$ sends to the server at the end of communication round $t$, given the initial model $z_{i}^{t,0} = z_{g}^{t}$.
Then we have 
\vspace{-5pt}
\begin{equation}
\small
        z_i^{t,E} = z_i^{t,E-m} (1 - 2 a_i \eta_t)^m - \eta_t b_i \sum_{j = 0}^{m-1} (1 - 2 a_i\eta_t)^j, \\
\end{equation}
where $\eta_t$ is the learning rate at the client-side in round $t$.

\noindent \textbf{Aggregation by FegAvg.} By aggregaring local models using FegAvg, we obtain the global model as
\begin{align}
        \nonumber 
\small
        Z^{t}_{FedAvg} &= \Bigg(\frac{2\phi_1 + \phi_3}{3}\Bigg)^t Z^{0}_{FedAvg} \nonumber\\
        & 
\small
        \nonumber- \Bigg(1-\bigg(\frac{2\phi_1 + \phi_3}{3}\bigg)^t\Bigg) \frac{\frac{b_1}{a_1}(1-\phi_1) + \frac{b_3}{a_3} \frac{1-\phi_3}{2}}{3 - (2\phi_1 + \phi_3)}.
\end{align}
\normalsize

\noindent \textbf{Aggregation by CADIS.} When using CADIS to aggregate the local models, we obtain the following global model
%\begin{small}
\begin{align}
        \nonumber 
\small
        Z^{t}_{CADIS} & \small = {\Bigg(\frac{\phi_1 + \phi_3}{2}\Bigg)}^t Z^{0}_{CADIS} \nonumber\\
        & \nonumber 
\small
        - \Bigg(1-\bigg(\frac{\phi_1 + \phi_3}{2}\bigg)^t\Bigg)  \frac{\frac{b_1(1-\phi_1)}{2a_1} + \frac{b_3(1-\phi_3)}{2a_3}}{2 - (\phi_1 + \phi_3)}.
\end{align}

where $\phi_i = (1 - 2a_i\eta_t)^K$.
When the models converge, we have 
\begin{align}
     & \nonumber 
\small
     Z_{FedAvg} = \lim_{t\to\infty}Z^{t}_{FedAvg} = \frac{\frac{b_1}{a_1}(1-\phi_1) + \frac{b_3}{a_3} \frac{1-\phi_3}{2}}{3 - (2\phi_1 + \phi_3)},\\
    & \nonumber 
\small
    Z_{CADIS} = \lim_{t\to\infty}Z^{t}_{CADIS} = \frac{\frac{b_1}{a_1}\frac{1-\phi_1}{2} + \frac{b_3(1-\phi_3)}{2a_3}}{2 - (\phi_1 + \phi_3)}.
\end{align}
By substituting the results above into (\ref{eq:global_optimal}) we obtain 
\begin{align}
\label{eqn:target}
\small
    \nonumber \mathcal{L}_{FedAvg} - \mathcal{L}_{CADIS}
    & = f^*(Z_{FedAvg}) - f^*(Z_{CADIS}) \\
    & \nonumber =   \frac{1}{8} \frac{b_1^2}{a_1 a_3} (a_1+a_3)^2 v_1 v_2,
\end{align}
where $v_1 = Q - P$ and $v_2 = \bigg(\frac{1}{a_1} + \frac{1}{a_3}\bigg)(Q+P) - \frac{2}{a_3}$; $P = \frac{\phi_1 - 1}{\phi_1 + \phi_3 - 2}, Q = \frac{2\phi_1 -2}{2\phi_1 + \phi_3 - 3}$. 
By proving $v_1 > 0$ and $v_2 > 0$, we deduce that $\mathcal{L}_{FedAvg} > \mathcal{L}_{CADIS}$. 
\section{Knowledge Distillation-based Regularization}
\label{subsec:regularization}

\begin{figure*}[tb]
    \centering
    \begin{minipage}{0.61\linewidth}
        \includegraphics[width=\linewidth]{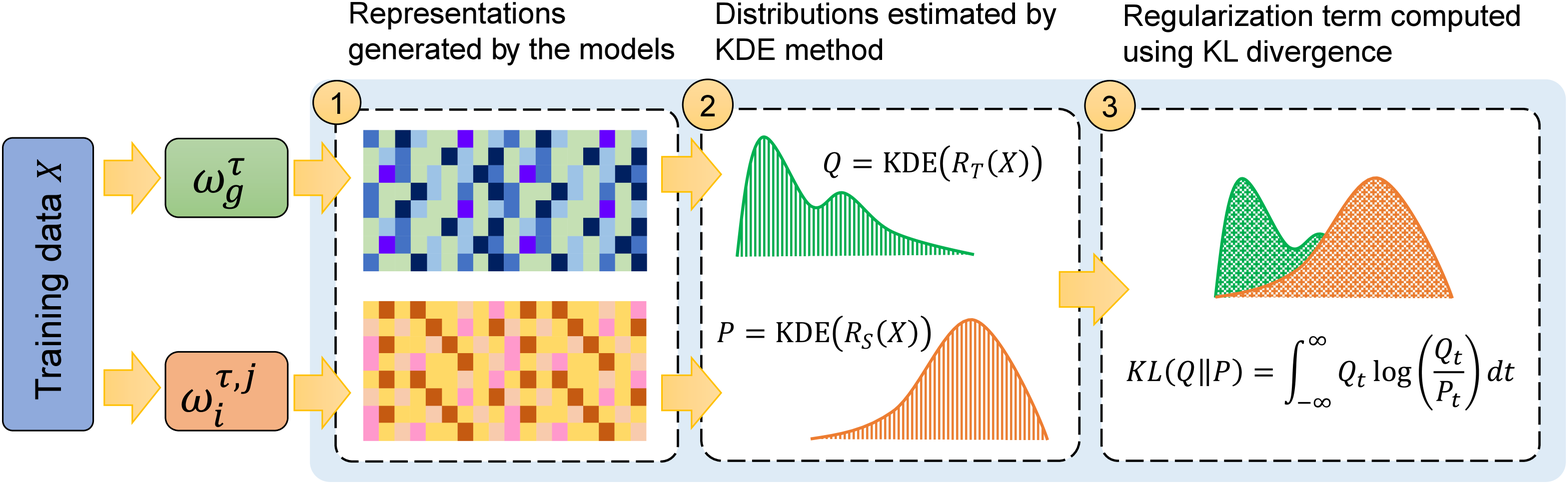}
        \caption{{\textbf{Illustration of a local training process at the client-side.} At every training epoch $e$, the client calculates a regularization term which is the KL divergence between the representation generated by locally trained model $\omega_i^{t,e}$ and that obtained from the global model $\omega_{g}^t$.\label{fig:local_training}}}
    \end{minipage}
    \hfill
    \begin{minipage}{0.35\linewidth}
        %\includegraphics[width=\linewidth]{fig/Overfitting.eps}
        %\caption{\textbf{Overfitting at local training.} The initial loss (blue dots) at the beginning of every round drops severely after a small number of epochs. \label{fig:overfitting}}
    \centering
    \subfigure[PILL]{
        \label{fig:dis_pill}
        \includegraphics[width=0.45\linewidth]{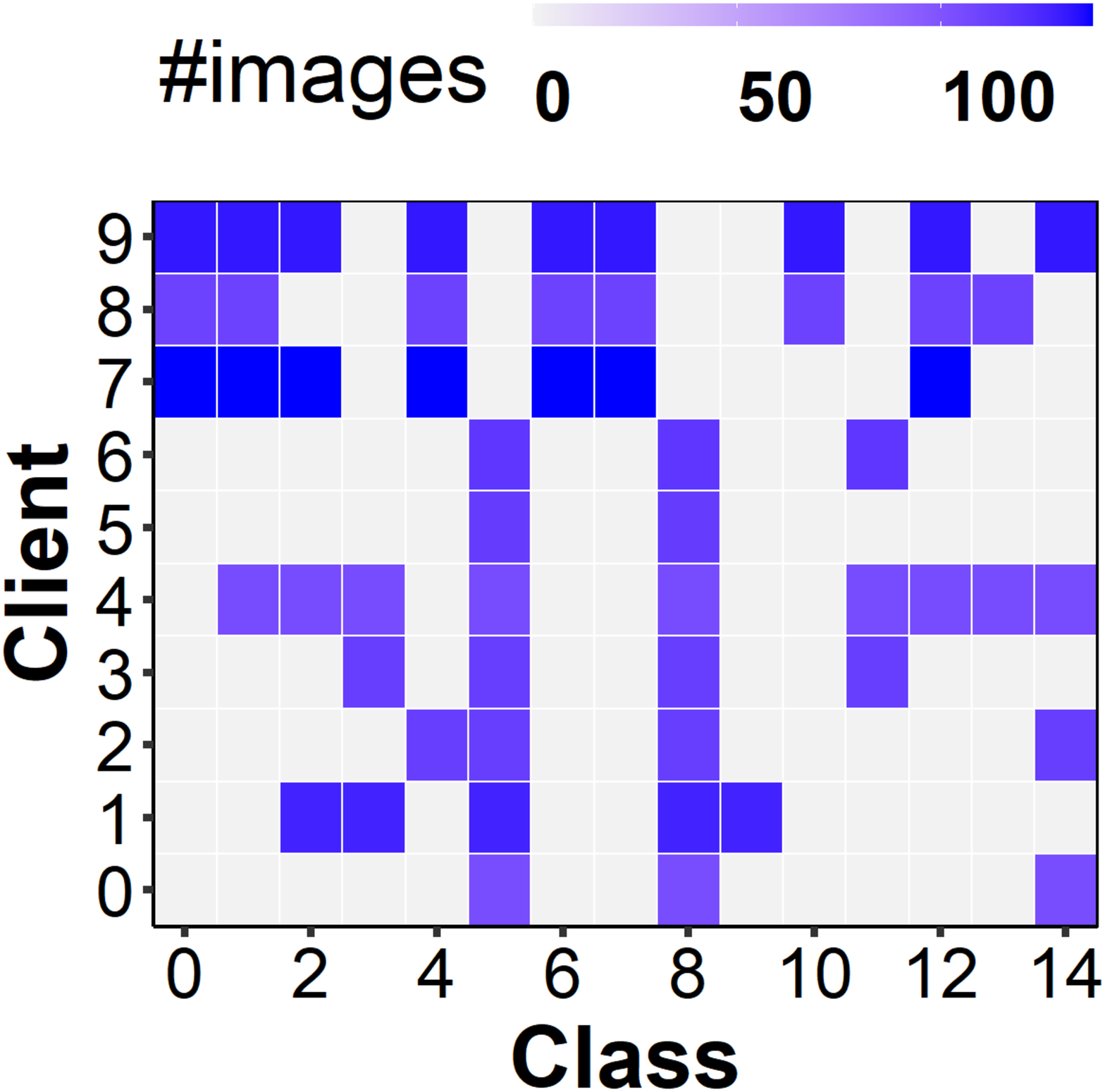}
     }
      \subfigure[CIFAR-10, MC]{
        \label{fig:dis_MC}
        \includegraphics[width=0.45\linewidth]{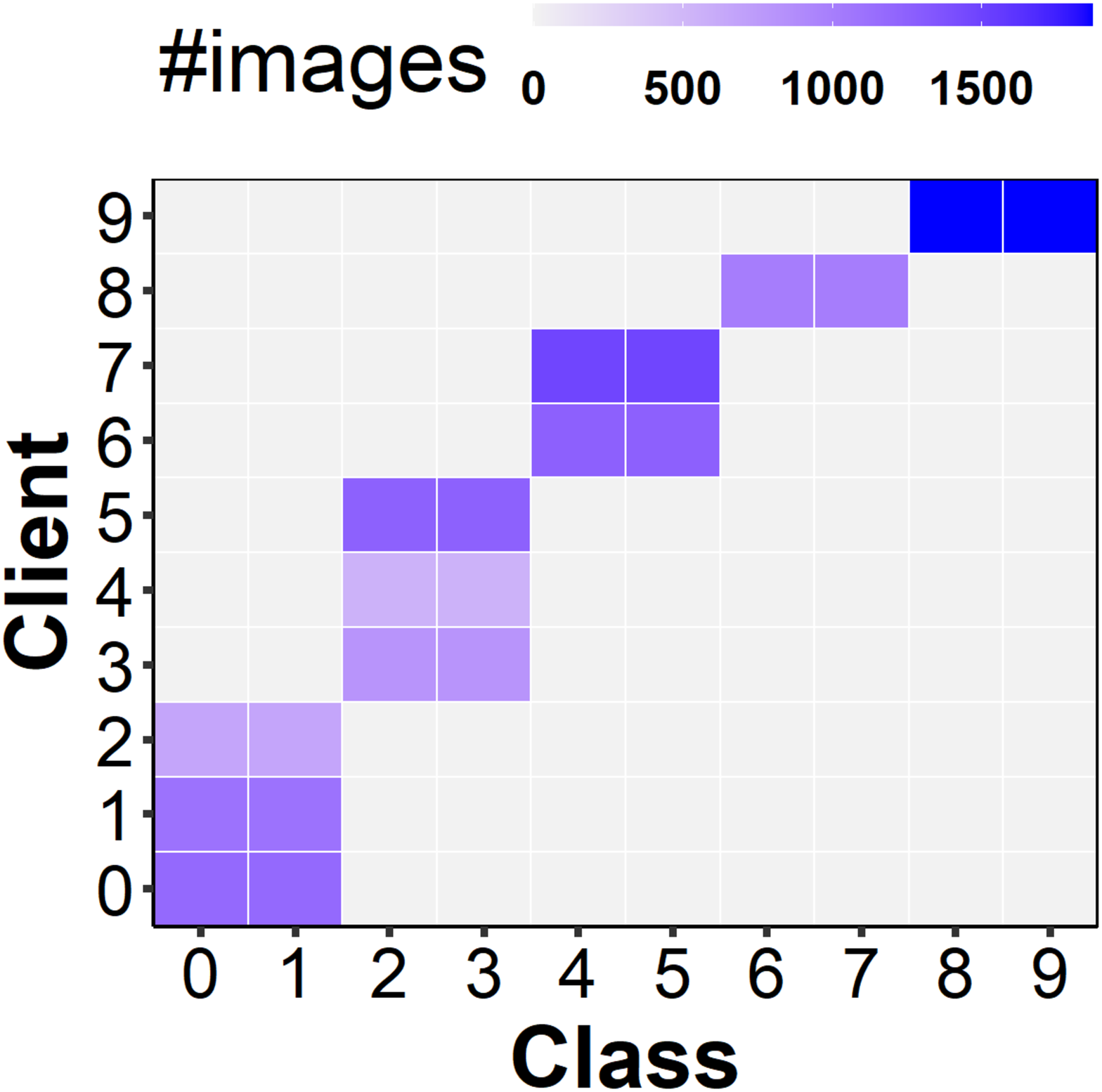}
     }
     \caption{Illustration of data distribution of CIFAR-10 dataset, and real-world PILL dataset in the case of 10 clients.}
     \label{fig:dataset_distribution}
    \end{minipage}
    %\vspace{-5pt}
\end{figure*}

\nguyen{We have proposed a clustering strategy to balance the bias on the inter-client level. However, when there are no clusters amongst the clients, the performance of CADIS returns to that of FedAvg, which is sensitive toward intra-client heterogeneity. Therefore, as an extension to our proposal, we integrate a subtle regularization into the local training process to diminish the effect of data heterogeneity.}
To this end, we design a regularization term inspired by the feature-based knowledge distillation technique \cite{Passalis_2018_ECCV}. 
This regularization term intuitively helps the clients to gain new knowledge from their local data without overwriting the previously learnt knowledge in the global model. As a result, the knowledge of the global model is accumulated throughout the federated training process.

We observe that the global model is an aggregation of multiple local models; as a result, it possesses more information and a higher generalizability.
Therefore, we use the global model delivered by the server at the beginning of each round as a teacher, while the clients' local models serve as students.
A client's regularization term is then defined by the Kullback-Leibler (KL) divergence between the representations generated by the client's local model and those obtained from the global model.
%Generally, we try to minimize the probabilistic distance between the representations of the clients and that of the server. 
Figure~\ref{fig:local_training} illustrates the flow for a client calculate the regularization term.
%The process to compute the regularization term is demonstrated in Figure~\ref{fig:local_training}.
The details are as follows.
% We observe two critical issues concerning the local training process at the client. 
% First, in many real cases, the number of samples possessed by each client is limitted. 
% In non-iid context, the 
% A serious problem occurs when clients train on considerably small or sparse-labelling data sets, is that the local models are prone to overfit quickly, which is potentially harmful for Server aggregation. To resolve this, we leverage a feature-based knowledge distillation loss that flexibly keeping the knowledge each client updates onto the model fairly close to that of the initial server model. 
Consider a client with the training dataset of $X$, let $R_S(X)$ be the representations generated by the locally trained model, and $R_T(X)$ be the representation produced by global model delivered by the server.
Instead of model the distribution of $R_S(X)$ and $R_T(X)$ directly, we try to model the pairwise interactions between their data samples, because as helps to describe the geometry of their respective feature spaces \cite{van2008visualizing}.
To accomplish this, we employ the joint probability density, which represents the likelihood that two data points are close together.
These joint density probability functions can be easily estimated using Kernel Density Estimation (KDE) \cite{scott2015multivariate}.
Let $\mathcal{P}$ and $\mathcal{Q}$ be the joint density probability functions corresponding to $R_S(X)$ and $R_T(X)$.
Suppose $p_{ij} \in \mathcal{P}$ denote the joint probability of $x_i$ and $x_j$ then $p_{ij}$ can be estimated using KDE as $p_{ij} = p_{i|j}p_j = \mathcal{K}_h(x_i, x_j)$, 
% \begin{equation}
% \small
%     p_{ij} = p_{i|j}p_j = \mathcal{K}_h(x_i, x_j), 
% \end{equation}
where $\mathcal{K}_h(x,x_i) = \mathcal{K}_G(x,x_i,h)$ is a Gaussian kernel, with $h$ is the bandwidth of the Gaussian bell. 
However, as stated in~\cite{passalis2018learning}, it is often impossible to learn a model that can accurately reproduce the entire geometry of a complex teacher model. Therefore, the conditional probability distribution of the samples can be used instead of the joint probability density function as follows 
\begin{equation}
\small
    p_{i|j} = \frac{\mathcal{K}_h(x_i, x_j)}
    {\sum_{k=1, k\neq j} \mathcal{K}_h(x_k, x_j)} \in [0,1].
\end{equation}
%The conditional probabilities are bounded to $[0, 1]$ and sum to $1$, i.e., $\sum_{i=1, i\neq j} p_{i|j} = 1$.
% \cite{passalis2018learning}, using the joint probability distribution to model the geometry of the data and perform knowledge transfer from the teacher to the student model can be inflexible. To overcome this issue, the joint probability function can be replaced with the conditional probability distribution of the samples and can be defined as:
% \begin{equation}
%     p_{i|j} = \frac{\mathcal{K}_h(x_i, x_j)}
%     {\sum_{k=1, k\neq j} \mathcal{K}_h(x_k, x_j)} \in [0,1].
% \end{equation}
% The conditional probabilities are bounded to $[0, 1]$ and sum to $1$, i.e., $\sum_{i=1, i\neq j} p_{i|j} = 1$.
The similar process is applied to estimate the probability distribution of the global model. 
Finally, we use Kullback-Leibler (KL) divergence to calculate the difference between the two distributions $\mathcal{P}$ and $\mathcal{Q}$ by using the following formula
% \begin{equation}
% \small
%     KL(\mathcal{Q} \parallel \mathcal{P}) = \int_{-\infty}^{\infty} 
%     \mathcal{Q}_t \times  \log \Bigg(\frac{\mathcal{Q}_t}{\mathcal{P}_t}\Bigg) dt.
% \end{equation}
% To this point, we have completed building the PDF for batch representation of the student $\mathcal{P}$ and that of the teacher $\mathcal{Q}$, we then can measure the divergence of the two distributions. Although there are several choices exist for the divergence metric, we use the well known Kullback-Leibler (KL) divergence:
% \begin{equation}
%     KL(\mathcal{Q}, \mathcal{P}) = \int_{-\infty}^{\infty} 
%     \mathcal{Q}_t log\Bigg(\frac{\mathcal{Q}_t}{\mathcal{P}_t}\Bigg) dt,
% \end{equation}
% where $\mathcal{Q}$ and $\mathcal{P}$ are the probability distributions of the teacher and student batch representation respectively. 
% Since each training dataset contains a finite number of samples, the regularization term can be represented by
\begin{equation}
\small
    \mathcal{L}_{KD} = KL(\mathcal{Q} \parallel \mathcal{P}) \approx \sum_{i=1}^{b} \sum_{j=1, j\neq i}^{b} q_{j|i} \times  \log \Bigg(\frac{q_{j|i}}{p_{j|i}}\Bigg),
\end{equation}
\noindent where $b$ is the batch size. 
Consequently, the final loss function for training at client $C_i$ in round $t$ is defined as
\small
\begin{align}
    \mathcal{L} &= \mathcal{L}_{CC} + \lambda\mathcal{L}_{KD} = \frac{1}{n_i} \sum_{e=1}^{E}\sum_{u = 1}^{{n_i}/{b}} \left \{ \left (\text{CE}\left (X_u, Y_u \right )|_{\omega_{i}^{t,e}} \right ) \right. \nonumber
    \\ 
    & + \left. \lambda 
    KL \left (\text{KDE}_{\omega_{g}^{t}}\left (X_u \right )\|\text{KDE}_{\omega_{i}^{t,e}}\left (X_u \right ) \right) \right \}.
    % (w_{k}^{(t,i)}, w^t)) 
    \label{eqn:final_loss}
\end{align}
\normalsize
Here $n_i$ is the cardinality of $C_i$'s dataset, $E$ is the number of training epochs;  $(X_u,Y_u)$ is training dataset of the $u$-th batch, where $X_u$ depicts the image set, and $Y_u$ denotes the corresponding labels; $\lambda \geq 0$ is the weighting factor of the regularization term.
% The pseudo-code to compute the Gaussian PDF is demonstrated in Algorithm~\ref{algo:pdf}.
% \input{algo/compute-pdf}

\section{Experiments and Results}
\label{sec:evaluation}
This section evaluates the performance of the proposed FL method, i.e., CADIS, against competing approaches for various FL scenarios. We show that CADIS is able to achieve higher performance and more stable convergence compared with state-of-the-art FL methods including FedAvg \cite{fedavg_mcmahan2017communication}, FedProx~\cite{fedprox_li2020federated}, FedDyn~\cite{feddyn_acar2021federated}, and FedFA~\cite{fedfa_huang2020fairness}, on various datasets and non-IID settings. In the following, we first introduce four image classification datasets used in our experiments including both standard conventional datasets and real-world medical imaging datasets. We also describe the setup for the experiments in Section~\ref{sec:eval_setting}. We then report and compare the performance of the CADIS with state-of-the-art methods using the top-1 accuracy on the test datasets. (Section~\ref{sec:eval_accuracy}). In all experiments, the SingleSet setting (centralized training at the server or training in the system of only one client)\footnote{Because Singleset trained on a single client (or server), it may be equivalent to training with an IID dataset.} is used as the reference.
%We also evaluate the SingleSet for being references (centralized training at server or training in the system of only one client)\footnote{Because Singleset trained on a single client (or server), it may be equivalent to training with an IID dataset.}. 
Finally, in Section~\ref{sec:eval_ablation}, we conduct ablation studies to highlight some key properties of CADIS.

\subsection{Datasets and Experimental Settings}
%  \begin{figure}[t]
%      \centering
%      \small
%      \subfigure[VAIPE]{
%         \label{fig:dis_pill}
%         \includegraphics[width=0.18\textwidth]{fig/pill_3.eps}
%      }
%       \subfigure[CIFAR-10, MC]{
%         \label{fig:dis_MC}
%         \includegraphics[width=0.18\textwidth]{fig/CIFAR10_10client_cluster.eps}
%      }
     
     %\subfigure[CIFAR-10, PA]{
     %   \label{fig:dis_PA}
     %   \includegraphics[width=0.18\textwidth]{fig/CIFAR10_10client_pareto.eps}
     %}
     % \subfigure[CIFAR-10, BC]{
     %   \label{fig:dis_BC}
     %   \includegraphics[width=0.18\textwidth]{fig/CIFAR10_10client_featured.eps}
     %}
      %\subfigure[CIFAR-10, UC]{
       % \label{fig:dis_UC}
    %    \includegraphics[width=0.18\textwidth]{fig/CIFAR10_10client_quantitative.eps}
     %}
%      \caption{Illustration of data distribution of CIFAR-10 dataset, and real-world VAIPE dataset in the case of 10 clients. In VAIPE, patients (clients) with identical diseases often take the same pills (classes).}
%      \label{fig:dataset_distribution}
%      \vspace{-10pt}
%  \end{figure}
\label{sec:eval_setting}
To evaluate the robustness of the proposed method in a real-world setting, we collect a large-scale real-world pill image classification dataset (due to the double-blind policy, we name our dataset as PILL). The dataset consists in total of $10,042$ images from $96$ patients, $276$ diagnoses and $94$ pills (classes). However, in our experiments, we use a subset of $10$ clients constituted $7$ clients diagnosed with diabetes and $3$ clients from other diseases. We then annotate the data of selected clients to be our evaluated sub-dataset~\footnote{We have to evaluate on sub-dataset because of the lack of manpower to annotate the whole dataset at the time of submission.}. The sub-dataset consists of $15$ classes, e.g., the pill name, and $7,084$ images. The dataset is then divided into two disjoint parts, where $90$\% of images are used for training and the rest $10$\% are used for testing.

To further evaluate the effectiveness of CADIS on bigger datasets, we use three benchmark imaging datasets with the same train/test sets as in previous work \cite{fedavg_mcmahan2017communication, fedprox_li2020federated, feddyn_acar2021federated, fedfa_huang2020fairness},  which are MNIST~\cite{Mnist}, CIFAR-10~\cite{Krizhevsky09}, and CIFAR-100~\cite{Krizhevsky09}.
We simulate data heterogeneity scenarios, i.e., cluster-skewed non-IID, by partitioning datasets and distributing the training samples among $n = 100$ clients. In this work, we target the sample-unbalanced multi-cluster (denoted as \textbf{MC})
non-IID, in which clients have the same label distribution belonging to the same cluster. We choose $5$ clusters with the ratio of clients in the clusters are $3:3:2:1:1$. 
The number of samples per client is unbalanced and each client has approximately $20\%$ of labels (classes), e.g., 2 classes for CIFAR-10. We also further consider different data partition methods in~Section\ref{sec:eval_ablation}.
Figure~\ref{fig:dataset_distribution} illustrates the class distribution of the PILL subset and CIFAR-10 dataset across clients with \textbf{MC} partition methods.

%\subsection{Implementation Details}
We train simple convolutional neural networks (CNNs) on MNIST as mentioned in~\cite{fedavg_mcmahan2017communication}. Specifically, we train ResNet-9~\cite{he2016deep} network on CIFAR-10, and CIFAR-100 dataset, and ResNet-18~\cite{he2016deep} on PILL dataset. For all the experiments, we use SGD (stochastic gradient descent) as the local optimizer. We also set the local epochs $E=5$, a learning rate of $0.001$, and a local batch size $b=8$.
We evaluate with the system of 100 clients like prior work in FL~\cite{fedavg_mcmahan2017communication, fedprox_li2020federated}.
The number of participating clients at each communication round is $k=3$ for PILL and $k=10 \to 50$ for other datasets. We also used the default hyper-parameters suggested by the original paper of each FL benchmark. Specifically, we set the proximal term $\mu=0.01$ for the FedProx method. For FedFA, we set $\alpha=1.0$ and $\beta=0$ as suggested in~\cite{fedfa_huang2020fairness}.
For FedDyn, we use $\alpha=0.5$~\cite{feddyn_acar2021federated}.

\subsection{Experimental Results}
\label{sec:eval_accuracy}
% \usepackage{booktabs}
% \usepackage{multirow}
%\usepackage[normalem]{ulem}
%\useunder{\uline}{\ul}{}

\begin{table}[t]
	\caption{Comparison of top-1 test accuracy to the benchmark approaches with \textbf{m}ulti-\textbf{c}luster non-IID.
	The values show the best accuracy that each FL method reaches during training$^{(*)}$} 
	\label{table:accuracy}
	\centering
	\scriptsize
	\setlength\tabcolsep{2pt} % default value: 6pt
	\resizebox{\linewidth}{!}{%
	\begin{threeparttable}
    \begin{tabular}{@{}lc|c|ccccc|cc@{}}
    \toprule
    \multirow{2}{*}{\textbf{Dataset}} 
    &\multirow{2}{*}{\begin{tabular}[c]{@{}c@{}} \textbf{\#Clients} \end{tabular}}
    & \multicolumn{6}{c}{\textbf{Top-1 Accuracy (\%)}}
    & \multicolumn{2}{c}{\textbf{\emph{Impr. (\%)}}}\\
    \cmidrule(lr){3-8}\cmidrule(lr){9-10}
    & &  \multicolumn{1}{c}{SingeSet} & \multicolumn{1}{c}{FedAvg} & \multicolumn{1}{c}{FedProx} & \multicolumn{1}{c}{FedFA}       
    &  \multicolumn{1}{c}{FedDyn} & \multicolumn{1}{c}{CADIS} & \multicolumn{1}{c}{(a)} & \multicolumn{1}{c}{(b)} \\ 
    \midrule
    PILL & 3/10 & 92.32 & \textbf{\textcolor{blue}{73.36}} & 70.04 & 6.67 & 22.17 & \textbf{\textcolor{red}{79.71}} & 8.7 & 8.7
    \\
    \midrule
    CIFAR-10 & 10/100 & 82.11 & \textbf{\textcolor{blue}{48.74}} & 48.43 & 10.00 & 19.62 & \textbf{\textcolor{red}{50.09}} & 2.8 & 2.8 
    \\
    \midrule
    \multirow{5}{*}{ CIFAR-100} & 10/100 & 52.36 & 33.42 & 33.21 & \textbf{\textcolor{blue}{36.49}} & 5.39 & \textbf{\textcolor{red}{38.09}} & 14.0 & 4.4
    \\
    & 20/100 & 52.36 & 32.70 & 32.13 & \textbf{\textcolor{blue}{35.28}} & 3.68 & \textbf{\textcolor{red}{38.16}} & 16.7 & 8.2
    \\
    & 30/100 & 52.36 & 32.07 & 32.37 & \textbf{\textcolor{blue}{35.42}} & 5.73 & \textbf{\textcolor{red}{38.55}} & 20.2 & 8.8 
    \\
    & 40/100 & 52.36 & 31.65 & 31.48 & \textbf{\textcolor{blue}{35.20}} & 6.30 & \textbf{\textcolor{red}{36.35}} & 14.8 & 3.3
    \\
    & 50/100 &  52.36 & 31.11 & 31.27 & \textbf{\textcolor{blue}{35.12}} & 6.59 & \textbf{\textcolor{red}{36.23}} & 16.5 & 3.2 
    \\
    \midrule
     \multirow{5}{*}{MNIST } & 10/100 & 99.15 & 93.04 & 92.91 & \textbf{\textcolor{blue}{93.33}} & 88.04 & \textbf{\textcolor{red}{93.45}} & 0.4 & 0.1 
    \\
    & 20/100 & 99.15 & 95.12 & 95.12 & \textbf{\textcolor{blue}{95.55}} & 93.58 & \textbf{\textcolor{red}{95.71}} & 0.6 & 0.2 
    \\
    & 30/100 & 99.15 & 95.06 & 95.02 & \textbf{\textcolor{blue}{95.50}} & 93.92 & \textbf{\textcolor{red}{96.00}} & 1.0 & 0.5 
    \\
    & 40/100 & 99.13 & 94.10 & 94.05 & \textbf{\textcolor{blue}{94.50}} & 88.04 & \textbf{\textcolor{red}{95.60}} & 1.6 & 1.2
    \\
    & 50/100 & 99.15 & 94.83 & 94.86 & \textbf{\textcolor{blue}{95.27}} & 93.55 & \textbf{\textcolor{red}{96.20}} & 1.4 & 1.0
    \\
    \bottomrule
    \end{tabular}
    \begin{tablenotes}
      \scriptsize
      \item (*) The best and second best results are highlighted in the \textbf{\textcolor{red}{bold-red}} and \textcolor{blue}{\textbf{bold-blue}}.\\
      (**) \emph{impr.(a)} and \emph{impr.(b)} are the relative accuracy improvement of CADIS compared with FedAvg and the best benchmark FL method (in percentage), respectively.
    \end{tablenotes}
    \end{threeparttable}
    }
    \vspace{-5pt}
\end{table}
\subsubsection{Top-1 accuracy} 
Table~\ref{table:accuracy} presents the results of the classification accuracy when comparing our proposed CADIS to the baseline methods on all the datasets with cluster-skewed non-IID. We report the best accuracy that each FL method reaches within $500$ communication rounds.
Specifically, CADIS achieves better accuracy than all other FL methods. 
For example, CADIS achieves an accuracy of $79.71$\% on the PILL dataset. This result significantly outperforms the best benchmark FL methods, e.g., FedAvg, by $8.7$\%. Compared to the second-best benchmark (marked in bold-blue text), our CADIS surpasses it by $1.35$\%, $1.60$\%, and $0.41$\% top-1 accuracy in CIFAR-10, CIFAR-100, and MNIST datasets, respectively ($k=10$ clients participating in each round). It is worth noting that the image classification tasks in  MNIST dataset are simple such as the accuracy of all the methods is asymptotic to that of the SingleSet, leading no room for optimizing. As a result, CADIS is only slightly better than the baseline methods.
This result emphasizes that our cluster-based weighted aggregation method can engage the clients from the `smaller' clusters (i.e., groups with a smaller cardinality) more effectively than the sample-quantity-based aggregation of FedAvg and the training-frequency-based aggregation of FedFA.
It demonstrates our theoretical finding mentioned in Proposition~\ref{pro:loss}.

%\subsubsection{Impact of the number of participating clients}
We next conduct a sensitivity study to quantify the impact of the number of participating clients $k$ per communication round on accuracy. As shown in Table~\ref{table:accuracy}, 
we change $k$ from $10$ up to $50$ when training on the CIFAR-100 and MNIST datasets.
We observe that varying the number of participating clients would slightly affect the top-1 accuracy but would not impact the relative result between CADIS and the baseline methods.
As the result, the improvement in accuracy of CADIS over the other two baseline methods is consistently maintained.

 % \begin{figure*}[htbp]
     % \centering
     % \small
     % \includegraphics[width=0.45\textwidth]{fig/legend.eps}
     % \\
     % \subfigure[MNIST]{
        % \label{fig:accuracy_mc_100_x_mnist}
        % \includegraphics[width=0.2\textwidth]{fig/mnist_C_N100_K10_avg10.eps}
        % \includegraphics[width=0.2\textwidth]{fig/mnist_C_N100_K20_avg10.eps}
        % \includegraphics[width=0.2\textwidth]{fig/mnist_C_N100_K30_avg10.eps}
        % \includegraphics[width=0.2\textwidth]{fig/mnist_C_N100_K40_avg10.eps}
        % \includegraphics[width=0.2\textwidth]{fig/mnist_C_N100_K50_avg10.eps}
     % }
     % \\
      % \subfigure[CIFAR-100]{
        % \label{fig:accuracy_mc_100_x_cifar100}
        % \includegraphics[width=0.2\textwidth]{fig/cifar100_C_N100_K10_avg10.eps}
        % \includegraphics[width=0.2\textwidth]{fig/cifar100_C_N100_K20_avg10.eps}
        % \includegraphics[width=0.2\textwidth]{fig/cifar100_C_N100_K30_avg10.eps}
        % \includegraphics[width=0.2\textwidth]{fig/cifar100_C_N100_K40_avg10.eps}
        % \includegraphics[width=0.2\textwidth]{fig/cifar100_C_N100_K50_avg10.eps}
     % }
     % \caption{Top-1 test accuracy of different FL methods vs. communication round. The results are plotted with the average-smoothed of every $10$ communication round to have a better visualization. We omit the result of CIFAR-10 due to the space limitation.}
     % \label{fig:accuracy_mc_100_x}
 % \end{figure*}
\subsubsection{Convergence analysis}
\begin{figure*}[tb]
    \centering
    \vspace{20pt}
    \includegraphics[width=0.45\linewidth]{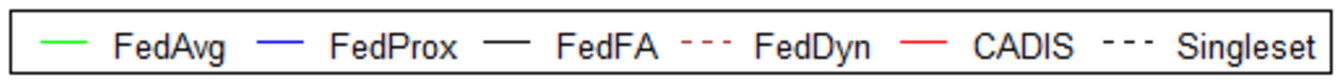}\\
    \subfigure[Top-1 accuracy]{
        \label{fig:stable_acc}
        \includegraphics[width=0.18\textwidth]{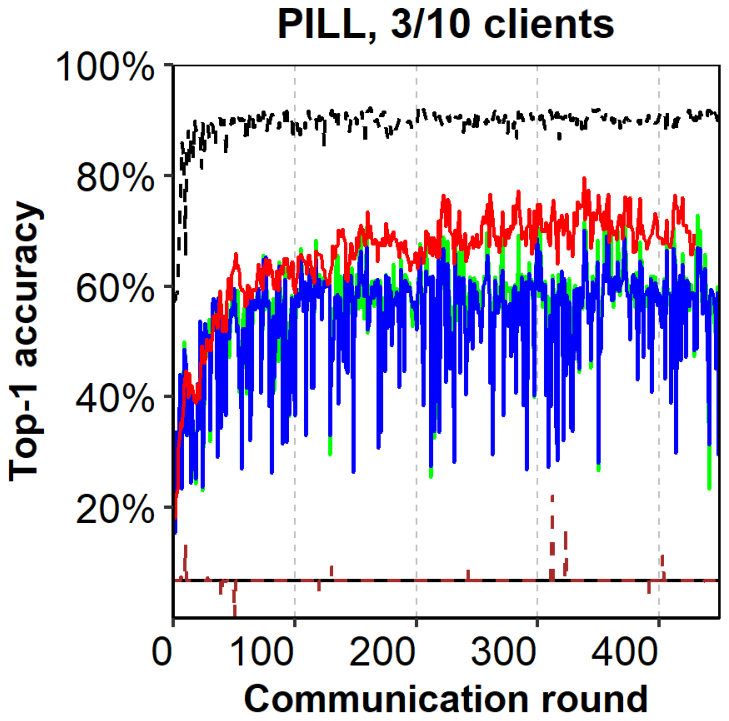}
        
        \includegraphics[width=0.18\textwidth]{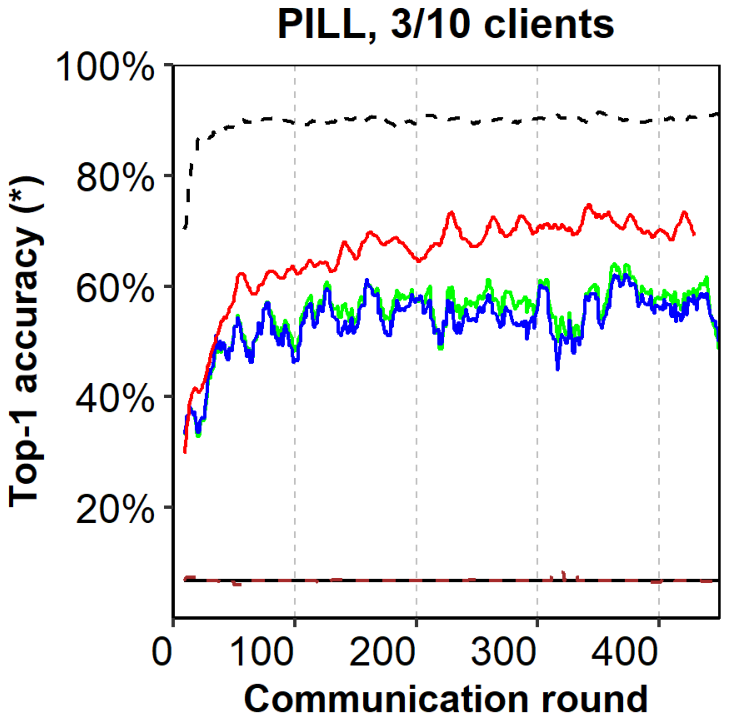}
    
        \includegraphics[width=0.18\textwidth]{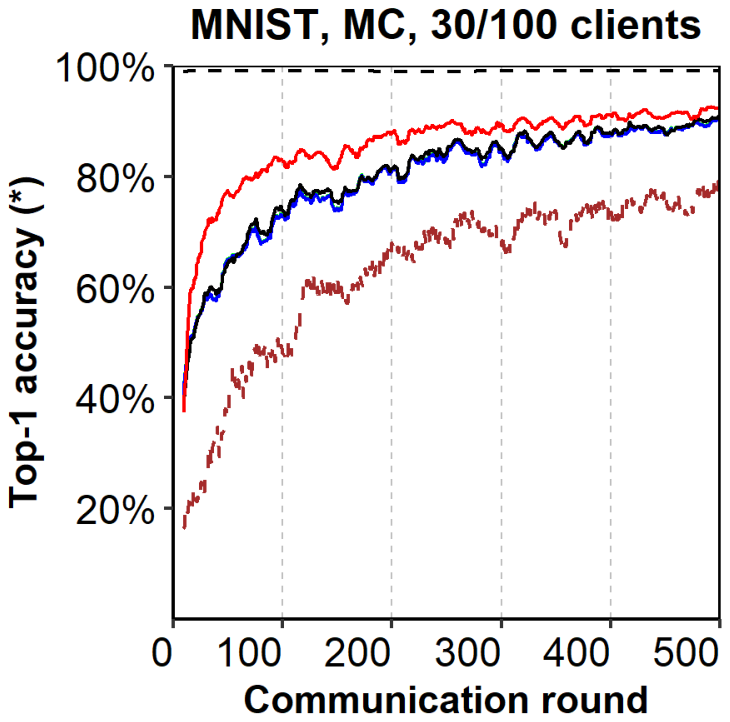}
    }
    \subfigure[Inference accuracy among all clients.]{
        \label{fig:stable_client}
        \includegraphics[width=0.18\textwidth]{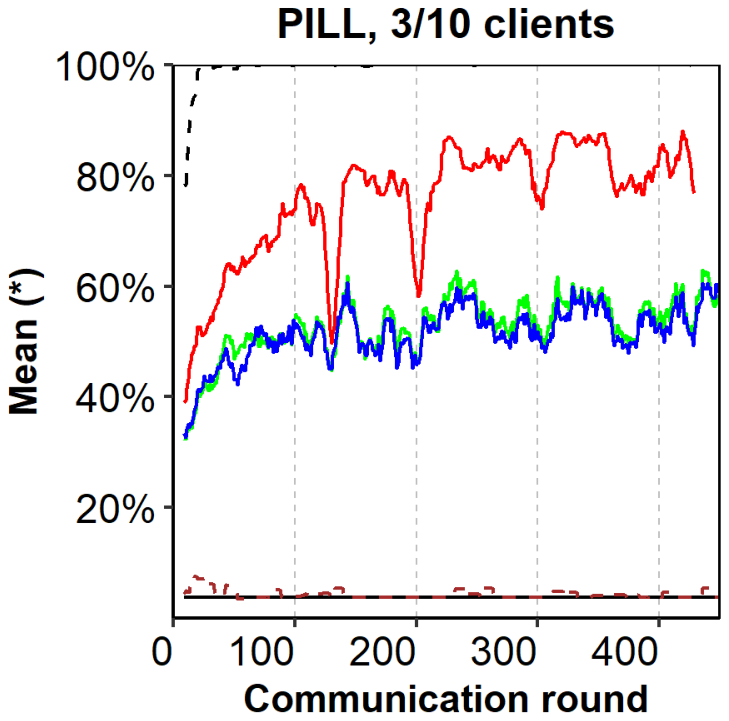}
    
        \includegraphics[width=0.18\textwidth]{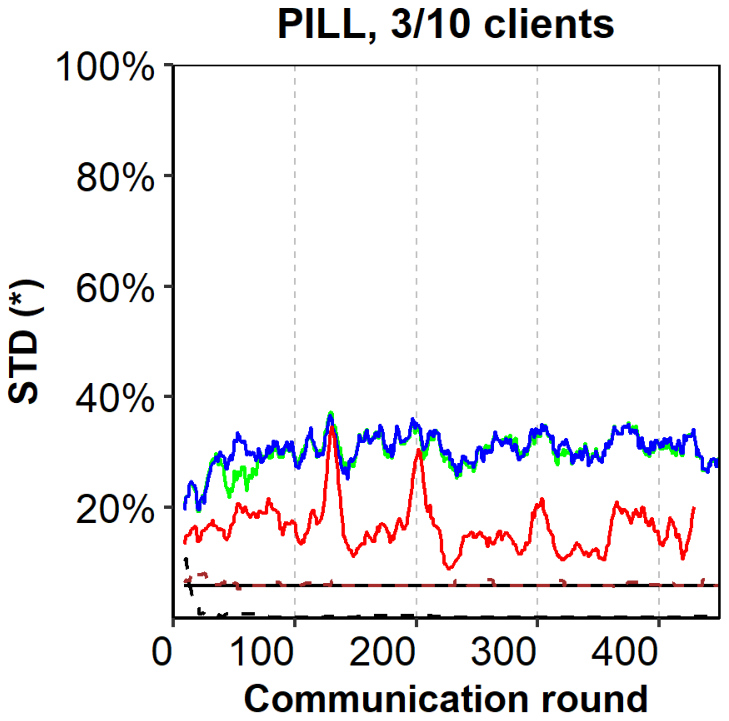}
    }
    \caption{Stability comparison of top-1 test accuracy (\%) and inference accuracy among all clients on the PILL, MNIST datasets. We omit the result of CIFAR-10 due to the space limitation. The results are plotted with the average-smoothed of every $10$ communication rounds to have a better visualization.}
    \label{fig:acc_vaipe}
    \vspace{-10pt}
\end{figure*}
To demonstrate the effectiveness of CADIS in reducing local computation at clients, we provide the number of communication rounds to reach a target top-1 accuracy and speedup relative to the FedAvg (Table~\ref{table:convergence}.
Overall, the convergence rate of CADIS is fastest in most of the evaluated cases except for the CIFAR-100 dataset. Specifically, to reach an accuracy of $60$\% for the PILL dataset, CADIS requires only $45$ communication rounds. FedAvg and FedProx spend $1.6\times$ longer than CADIS. In addition, CADIS is equivalent to FedFA in the case of MNIST dataset, while it is slower than FedFA in the case of CIFAR-100. It is because CADIS requires some first communication rounds for the similarity matrix coverage (Theorem~\ref{pro:similarity}), which leads to an incorrect clustering and slow down its coveragence rate. However, it is worth noting that CADIS achieves higher top-1 accuracy than FedFA when converged.
% \usepackage{booktabs}
% \usepackage{multirow}
%\usepackage[normalem]{ulem}
%\useunder{\uline}{\ul}{}

\begin{table}[t]
    \caption{Number of communication rounds required to reach a target Top-1 accuracy and speedup relative to FedAvg.} 
	\label{table:convergence}
	\centering
	\setlength\tabcolsep{2pt} % default value: 6pt
	\resizebox{\linewidth}{!}{%
	\begin{threeparttable}
    \begin{tabular}{@{}lc|lllll@{}}
    \toprule
    \textbf{Dataset}
    &\textbf{Acc.}
    & \multicolumn{1}{l}{FedAvg} & \multicolumn{1}{l}{FedProx} & \multicolumn{1}{l}{FedFA}       
    &  \multicolumn{1}{l}{FedDyn} & \multicolumn{1}{l}{CADIS} \\ 
    \midrule
    \multirow{2}{*}{PILL} & 60\% & 74 & 74 (\texttt{1.0$\times$}) & N/A & N/A & \textbf{\textcolor{red}{45 (\texttt{1.6$\times$})}} \\
    & 70\% & 159 & 338 (\texttt{0.5$\times$}) & N/A & N/A & \textbf{\textcolor{red}{136 (\texttt{1.2$\times$})}} 
    \\
    \midrule
    \multirow{2}{*}{MNIST } & 80\% & \textbf{\textcolor{red}{94}} & \textbf{\textcolor{red}{94 (\texttt{1.0$\times$})}} & \textbf{\textcolor{red}{94 (\texttt{1.0$\times$})}} & 279 (\texttt{0.3$\times$}) & \textbf{\textcolor{red}{94 (\texttt{1.0$\times$})}}\\
     & 90\% & 341 & 341 (\texttt{1.0$\times$}) & \textbf{\textcolor{red}{281 (\texttt{1.2$\times$})}} & N/A & \textbf{\textcolor{red}{281 (\texttt{1.2$\times$})}}
    \\
    \midrule
    CIFAR-10 & 48\% & 960 & 960 (\texttt{1.0$\times$}) & N/A & N/A & \textbf{\textcolor{red}{537 (\texttt{1.8$\times$})}}  
    \\
    \midrule
    %\multirow{2}{*}{CIFAR-100} & 30\% & 179 & 221 (\texttt{0.8$\times$}) & \textbf{\textcolor{red}{122 (\texttt{1.5$\times$})}} & N/A & 327 (\texttt{0.5$\times$})
    %\\
    CIFAR-100 & 35\% & N/A & N/A & \textbf{\textcolor{red}{281}} & N/A & 644
    \\
    \bottomrule
    \end{tabular}
    \begin{tablenotes}
      \scriptsize
      \item (*) The symbol N/A means the method is not able to reach the target test accuracy.
    \end{tablenotes}
    \end{threeparttable}
    }
    \vspace{-5pt}
\end{table}

\subsection{Ablation studies}
\label{sec:eval_ablation}
\subsubsection{Robustness to the client datasets}
In the previous subsection, we focus on the top-1 testing accuracy on an IID test dataset to estimate the goodness of the trained model over the global distribution. Because only a small portion of clients participate in the training process at each communication round, the aggregated global model at the server overly fits the sub-dataset of some clients, e.g., most-recently trained clients, or clients in the same cluster. 
To estimate the robustness of an FL method against clients, we consider the trend of top-1 accuracy by estimating the average accuracy obtained in the last $10$ communication rounds (named $10$\emph{-round averaging accuracy} for short). As shown in Figure~\ref{fig:stable_acc} (Left), the difference in the top-1 accuracy between two communication rounds of FedAvg and FedProx is non-trivial. The top-1 accuracy of CADIS oscillates with a smaller amplitude than those of FedAvg and FedProx. As the result, there is a clear gap of $10$-round averaging accuracy between CADIS and the FedAvg as shown in Figure~\ref{fig:stable_acc} (Middle). Another interesting point is that although CADIS is equivalent to another baseline method for the MNIST dataset in terms of top-1 accuracy, the $10$-round averaging accuracy of CADIS outperforms those of the baseline significantly, e.g., $95.3$\% as in CADIS versus around $94.2$\% of FedAvg (Figure~\ref{fig:stable_acc} (Right). 
We also test the global model on the local sub-dataset of all the participating clients at the beginning of each communication round, i.e., do the inference pass at clients.
The result in Figure~\ref{fig:stable_client} shows
that CADIS has consistently higher average inference accuracy across clients with smaller variances than the baselines. 

The results imply that the aggregated global model obtained by CADIS is more stable than the others and does not overfit clients' sub-datasets. It is expected because CADIS is designed with Knowledge Distillation-based Regularization for clients to avoid the local overfitting issue. Thus, we state that our CADIS could learn a well-balanced model between clients.

\subsubsection{Impact of the non-IID type}
% \usepackage{booktabs}
% \usepackage{multirow}
%\usepackage[normalem]{ulem}
%\useunder{\uline}{\ul}{}

\begin{table}[t]
	\caption{\small
	Comparison of top-1 test accuracy to the benchmarks with different partitioning methods, i.e., PA, BC, UC.}
	\label{table:accuracy_other}
	\centering
	\scriptsize
	\setlength\tabcolsep{3pt} % default value: 6pt
	\resizebox{\linewidth}{!}{%
	\begin{threeparttable}
    \begin{tabular}{@{}l|ccc|ccc|ccc@{}}
    \toprule
    \multirow{2}{*}{\begin{tabular}[c]{@{}c@{}}Partitioning \\method\end{tabular}}
    & \multicolumn{3}{c}{\textbf{MNIST}}
    & \multicolumn{3}{c}{\textbf{CIFAR-10}}
    & \multicolumn{3}{c}{\textbf{CIFAR-100}}\\
    \cmidrule(lr){2-4}\cmidrule(lr){5-7}\cmidrule(lr){8-10}
    &  \multicolumn{1}{c}{PA} & \multicolumn{1}{c}{BC} & \multicolumn{1}{c}{UC} 
    &  \multicolumn{1}{c}{PA} & \multicolumn{1}{c}{BC} & \multicolumn{1}{c}{UC} 
    &  \multicolumn{1}{c}{PA} & \multicolumn{1}{c}{BC} & \multicolumn{1}{c}{UC}  \\ 
    \midrule
    SingeSet 
    & 99.22 & 99.24 & 99.18 %& 99.15
    & 81.37& 39.20& 84.75 %& 82.11
    & 80.85& 52.66& 52.62 %& 52.36
    \\
    \cmidrule(lr){2-10}
    FedAvg   
    & \textbf{\textcolor{red}{96.89}} & \textbf{\textcolor{blue}{96.69}} & 96.02 %& 93.04
    & 62.26& \textbf{\textcolor{blue}{20.22}} & 43.45 %& 48.74
    & \textbf{\textcolor{blue}{21.01}} & 34.96& 34.16 %& 33.42
    \\
    FedProx  
    & \textbf{\textcolor{blue}{96.73}} & 96.66 & 96.01 %& 92.91
    & 62.63& 20.18& \textbf{\textcolor{blue}{43.86}} %& 48.43
    & 20.76& 34.61& 34.06 %& 33.21
    \\
    FedFA
    & 96.50 & 96.67 & \textbf{\textcolor{blue}{96.29}} %& 93.33
    & \textbf{\textcolor{red}{66.41}}& 10.00& 10.00 %& 10.00
    & 1.00& \textbf{\textcolor{red}{37.18}}& \textbf{\textcolor{red}{37.38}} %& 36.49
    \\
    FedDyn
    & 92.79 & 93.36 & 91.72 %& 88.04
    & 18.79& 19.34& 19.28 %& 19.62
    & 2.14& 9.69& 13.75 %& 5.39 
    \\
    CADIS
    & \textbf{\textcolor{red}{96.89}} %96.58 
    & \textbf{\textcolor{red}{96.74}} 
    & \textbf{\textcolor{red}{96.58}} 
    %& \textbf{\underline{93.45}}
    & \textbf{\textcolor{blue}{62.93}} 
    & \textbf{\textcolor{red}{34.83}}
    & \textbf{\textcolor{red}{47.42}} %& 50.09%
    & \textbf{\textcolor{red}{21.52}} 
    & \textbf{\textcolor{blue}{34.96}} %34.6
    & \textbf{\textcolor{blue}{34.62}} %& 38.09
    \\
    %\cmidrule(lr){3-14}
    %& \emph{impr.(a)} 
    %& -0.37 & 0.05 & -0.28 & 0.13
    %& - & - & - & -
    %& - & - & - & - \\
    %%& - \\
    %& \emph{impr.(b)} 
    %& 4.03 & 3.62 & 4.69 & 6.14
    %    & - & - & - & -
    %& - & - & - & - \\
    \bottomrule
    \end{tabular}
    \begin{tablenotes}
     \scriptsize
      \item (*) The best and second best results are highlighted in the \textbf{\textcolor{red}{red}} and \textcolor{blue}{\textbf{blue}}.
    \end{tablenotes}
    \end{threeparttable}
    }
    \vspace{-5pt}
\end{table}
We study the robustness of our method with the different types of non-IID by considering both conventional label distribution skew (Pareto), and other patterns of cluster skew~\cite{fedDRL} 
\nguyen{($N=100$ and $k=10$).}
\begin{itemize}
    \item Pareto (denoted as \textbf{PA}): The number of images of a class among clients following a power law~\cite{li2019convergence, fedprox_li2020federated}. 
    \item Sample-balanced single cluster (denoted as \textbf{BC}): a simple case of cluster skew with only one cluster and the number of samples per client does not change among clients. To measure the bias of the proposed model toward the cluster, we choose the number of clients inside the cluster significantly higher than the others, e.g, 60\%.
    %Featured
    \item Sample-unbalanced single cluster (denoted as \textbf{UC}): Similar to the BC but the number of samples per client is unbalanced. %To this end, we sample data capacity of each client using a Gaussian distribution with the variance no more than a fifth of the mean.
\end{itemize}
In Section~\ref{sec:eval_accuracy}, we showed numerical data regarding the MC distribution, in which CADIS demonstrated an improvement in accuracy compared to other methods. A similar observation was observed in the PA, BC, and UC data distribution (Table~\ref{table:accuracy_other}). 
CADIS achieves the best top-1 accuracy in most of the experiments (the second-best top-1 in the remaining).  
For example, CADIS improves the top-1 accuracy by $1.7\times$ and $1.1\times$ in the case of the CIFAR-10 dataset, BC and UC, respectively. The result implies that CADIS has a good performance with different types of cluster-skewed non-IID while achieving acceptable performance with label distribution non-IID (equivalent to FedAvg).

\subsection{Discussion}
\subsubsection{Impact of the transitive learning}
\begin{figure}[tb]
\begin{minipage}{0.44\linewidth}
    \centering
%\begin{figure}[h]
    %\centering
    \includegraphics[width=1\linewidth]{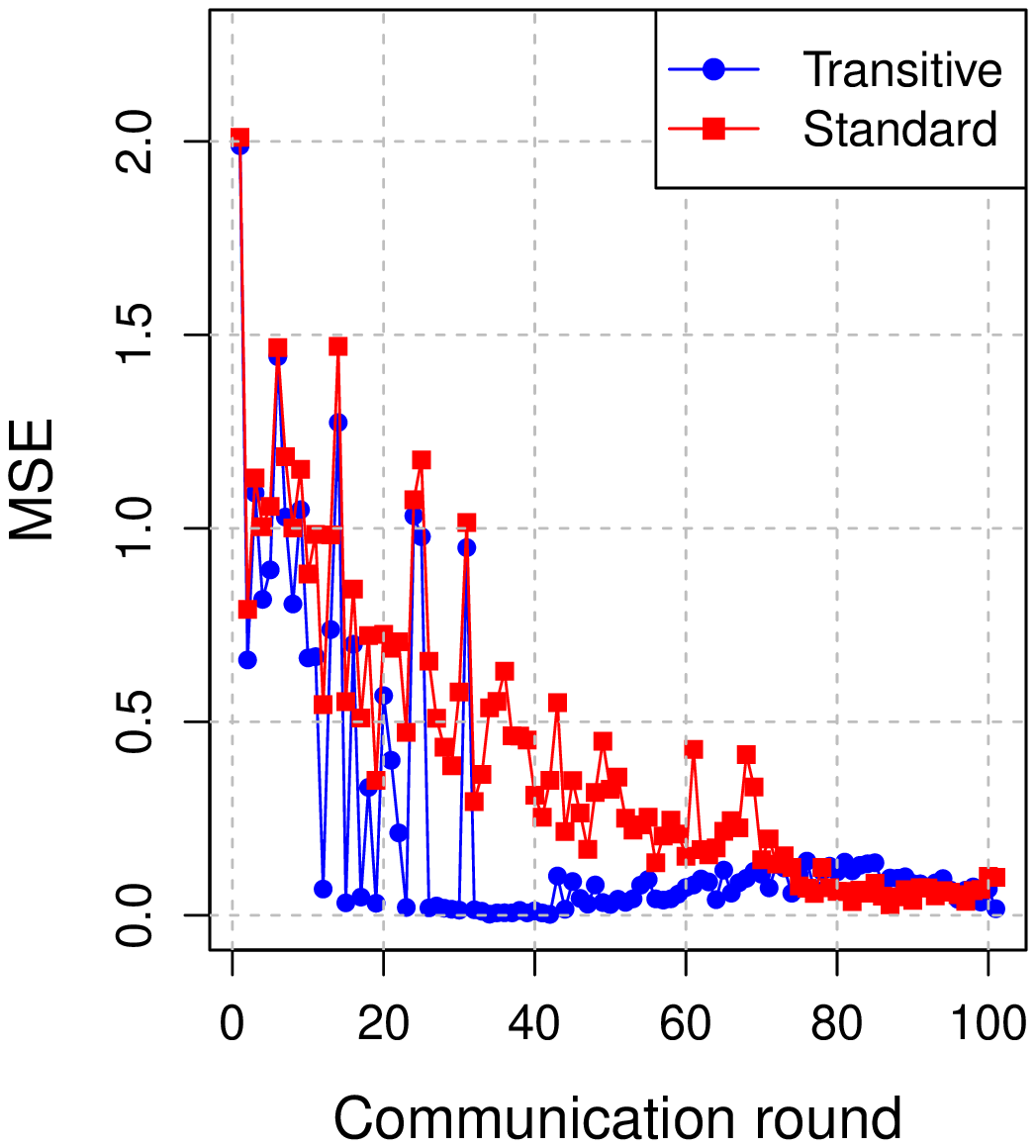}
    \caption{Similarity matrix convergence speed of transitive learning.}
    \label{fig:transitive}
    %\vspace{-10pt}
%\end{figure}
\end{minipage}
\hspace{0.1cm}
\begin{minipage}{0.52\linewidth}
        \includegraphics[width=1\linewidth]{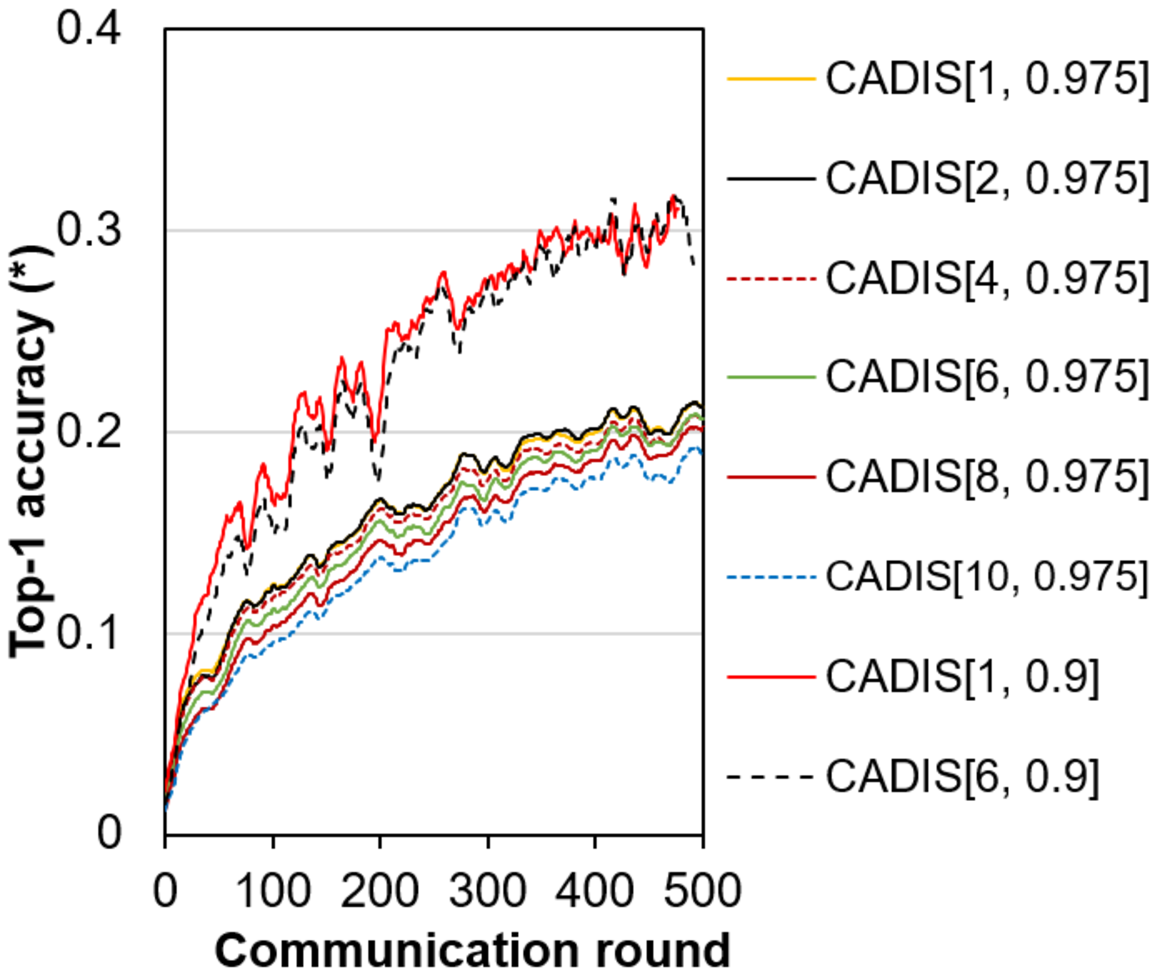}\\
        \vspace{-5pt}
    \caption{Impact of the hyper-parameter. CADIS[factor $\lambda$, threshold $\epsilon$]}
    \label{fig:tunning}
\end{minipage}
\vspace{-5pt}
\end{figure}
We introduced transitive learning in Section~\ref{subsec:transitive} to speed up the convergence of the similarity matrix. We confirm that CADIS with transitive learning (Transitive) and without transitive learning (Standard) could reach the same accuracy in our experiment.  However, transitive learning clusters the clients into groups faster than Standard. Figure~\ref{fig:transitive} shows the MSE distance of the similarity matrix built up by two methods with the correct similarity matrix (ideal one). Transitive could coverage after $40$ communication rounds while Standard needs approximate $100$ rounds.

\subsubsection{Impact of hyper-parameters}
In the experiments shown in Section~\ref{sec:eval_accuracy}, we tune the similarity threshold $\gamma$ and the factor $\lambda$ and report the best result obtained. In this section, we discuss how the hyper-parameters impact the top-1 accuracy. Figure~\ref{fig:tunning} show the results when we change both hyper-parameters of CADIS on the CIFAR-100 dataset, \textbf{MC} distribution. The result shows that both two hyper-parameter could lead to a change in accuracy. However, CADIS is much more sensitive to the similarity threshold $\epsilon$. For example, CADIS[$1$, $0.975$] achieve $26.6$\% while CADIS[$1$, $0.9$] reaches $32.1$\%. In our experiment, the best similarity threshold also changes when we change the dataset, e.g., $0.975$ for MNIST and $0.9$ for CIFAR-10 and CIFAR-100.

\subsubsection{The generalization of proposition \ref{prop:pen_shift}}
\nguyen{It is worth noting that the proof of proposition \ref{prop:pen_shift} can be used regardless of the condition $R\geq0$. In the case where the values of $R$ are not confined to the non-negative domain, one may deduce that the updates among rows of the penultimate layer will exhibit an inverse trend, depending on the label being trained. However, since our objective is to discover the labels underlying the training dataset, we only analyze the scenario in which the representation vector $R$ is monotonically non-negative.}

\subsubsection{Computational Overhead}
\begin{figure}[t]
    \centering
    \includegraphics[angle=270,width=0.8\linewidth]{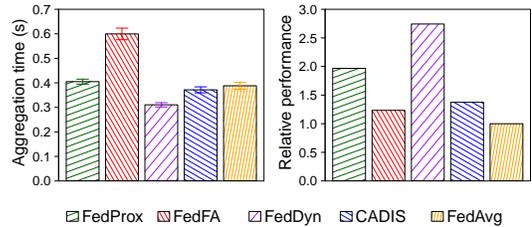}
    %{fig/server_computation_time.eps}
    \caption{Average aggregation time at server Left) and relative performance (samples/s) of local training at client normalized to those of FedAvg (Right) over $50$ communication rounds.}
    \label{fig:server_computation_time}
    \vspace{-5pt}
\end{figure}

\nguyen{We now estimate the computational overhead of the aggregation at the server of CADIS in comparison with FedAvg and the other methods.
The result in Fig.~\ref{fig:server_computation_time} (Left) shows that the computation overhead at the server of the CADIS's clustering module is trivial, i.e., the time of CADIS is approximate those of FedAvg and FedProx. This is an expected result because CADIS clusters the clients based on the information of the penultimate layer whose size is quite small, e.g., $256\times100$ in the case of the ResNet-9 model and CIFAR-100 dataset. 
}

\nguyen{
For the computation overhead of local training at client, we estimate the relative performance (on average) of CADIS over those one of FedAvg using the same device setting, e.g., the GPU Force-GTX 3090. The result in Fig.~\ref{fig:server_computation_time} (Right) shows 
CADIS require $1.37\times$ more computation at local than FedAvg (for performing the knowledge distillation regularization).
}

\section{Related Works}
\label{sec:related_work}
\nguyen{To tackle the statistical heterogeneity, i.e., non-IID, problem, many efforts focused on 
designing weighting strategies for aggregation at server~\cite{ HUANG2022170, fedfv_wang2021federated, fedadp}.}
The authors in \cite{HUANG2022170} developed a weighed aggregation mechanism, in which the weight of a client's local model is the sum of the information entropy calculated from the accuracy of the local model and the number of the client has participated in the training.
In \cite{fedfv_wang2021federated}, Wang et al., focused on the internal and externa conflict between the clients. 
%Specifically, the authors define two types of conflicts: internal and external. 
The former indicates the unfairness among clients selected in the same round, whereas the latter represents a conflict between the assumed gradient of a client who has not been chosen and the global update. 
In order to accomplish this, they proposed a mechanism to eliminate conflicts before averaging the gradients. 
\nguyen{Alternatively, many studies improve the training algorithm at client side~\cite{fedprox_li2020federated, feddyn_acar2021federated, karimireddy2020scaffold, fednova}.} In \cite{fedprox_li2020federated}, the authors addressed data heterogeneity by adding a so-called proximal term to the loss function, which restricts local updates to be closer to the initial (global) model.
The authors in \cite{feddyn_acar2021federated} used an adaptive regularization term leveraging the cumulative gradients when training the local models.
In \cite{karimireddy2020scaffold}, the authors investigated how to remedy the client drift induced by the heterogeneous data among clients in their local updates. 

\nguyen{However, previous studies specifically take into account the label skew non-IID when each client has a fixed number of classes (label size imbalance ~\cite{fedavg_mcmahan2017communication,fedprox_li2020federated,fedat,ICPP2021_fedCav,ijcai2021_Fairness_fedfv,xiao2020averaging}) or when the number of samples of a certain class is distributed to clients using the power-law or Dirichlet distribution (label distribution imbalance~\cite{li2019convergence,IWQOS2021_fedacs,fedfa_huang2020fairness,fednova}). Recently, some works consider the non-IID scenario which is more close to the real-world data such as the numbers of classes are often highly imbalanced~\cite{BalanceFL2022}, or following the cluster-skew non-IID distribution~\cite{icml2020_hsieh20a_skewscout,fedDRL}. Especially, cluster-skew is firstly introduced by~\cite{icml2020_hsieh20a_skewscout} where there exists a data correlation between clients. Authors in~\cite{fedDRL} tackle this data distribution by adaptively assigning the weights for clients at aggregation by using Deep Reinforcement Learning.
This work also focuses on \textbf{cluster-skew}. Unlike ~\cite{fedDRL}, we combine both the aggregation optimization approach (clustered aggregation) at the server side and the training enhancement approach at clients (knowledge distillation-based regularization) in this work.}

\nguyen{\textbf{Cluster-based Federated Learning:} Recent works cluster the clients into groups where different groups of clients have different learning tasks~\cite{NEURIPS2020_e32cc80b, CFL2021}, or different computation/network resource~\cite{fedat,FedFast}.
Other methods have been proposed to identify adversarial
clients and remove them from the aggregation~\cite{Sattler2020,CFL2021} based on their cosine similarities. 
Recently~\cite{FedCluster,briggs2020federated} proposed to use clustering to address the non-IID issue. It is worth noting that most of the previous cluster-based Federated Learning methods assume that all clients will participate in the clustering process or use the whole model for clustering which is unpractical in a real Federated Learning system.
Our proposed CADIS can effectively cluster clients by using the information of the penultimate layer only.
}

\section{Conclusion}
\label{sec:Conclusion}
\label{sec:conclusion}
In this paper, we introduced for the first time a new type of non-IID data called cluster-skewed non-IID, in which clients can be grouped into distinct clusters with similar data distributions. We then provided a metric that quantifies the similarity between two clients' data distributions without violating their privacy, and then employed a novel aggregation scheme that guarantees equality between clusters.
Moreover, we designed a local training regularization based on the knowledge-distillation technique that reduces the impact of overfitting on the clients' training process and dramatically boosts the trained model's performance. 
We performed the theoretical analysis to give the basis of our proposal and proved its superiority against a benchmark. 
%To this end, an aggregation scheme that guarantees equality between clusters has been proposed. 
Extensive experimental results on both standard public datasets and our own collected real pill image dataset demonstrated that our proposed method, CADIS, outperforms state-of-the-art. Notably, in the cluster-skewed scenario, our proposed FL framework, CADIS, improved top-1 accuracy by $16$\% compared to FegAvg and by up to $8.7$\% concerning other state-of-the-art approaches.
%Although achieving a substantial performance advantage for clustered-skewed non-IID data, the performance gaps between CADIS and existing methods for some other non-IID data types remain marginal.
%Therefore, our future efforts will focus on enhancing the solution's robustness on non-IID data in general.
%\textcolor{blue}{[Add here one sentence about the limitations and another one about the future work] The current approach still ... . } In conclusion, our study provides a robust approach for handling cluster-skewed non-IID data in federated learning. 

\section{Acknowledgments}
This work was funded by Vingroup Joint Stock Company (Vingroup JSC),Vingroup, and supported by Vingroup Innovation Foundation (VINIF) under project code VINIF.2021.DA00128. This work was supported by JSPS KAKENHI under Grant Number JP21K17751 and is based on results obtained from a project, JPNP20006, commissioned by the New Energy and Industrial Technology Development Organization (NEDO).\\ 

\balance
\bibliographystyle{IEEEtran}
\bibliography{main}
\newpage
%\appendix
\appendices
\section{Artifact Description}
\textbf{BADGE APPLICATION}: Open Research Objects (ORO)

\subsection{SUMMARY OF EXPERIMENTAL SETTINGS}

Our codes has been published at Artifact 1, which is a modified framework for federated learning introduced in \cite{wang2021federated}. The original framework can be accessed at Artifact 2. To reproduce our experiments, simply follow the instructions shown in the Readme file. Below are detailed experimental and hyperparameter settings.

At clients, we apply Stochastic Gradient Descent (SGD) as the optimizer with the learning rate $0.001$, batch size $8$ and local epochs $5$ if not explicitly say otherwise. The knowledge distillation-based loss is scaled with the factor $\lambda$, which is set $1$ during experiments, if not explicitly say otherwise. The number of total clients varies from $10$ to $100$ whereas the participation ratio every round increases from $10\%$ to $50\%$. Regarding the server, the similarity-based clustering threshold $\epsilon$ is varied regarding different datasets. In details, we found that $\epsilon = 0.975$ worked best for MNIST, $0.95$ for CIFAR10 and $0.9$ CIFAR100. 

\subsection{HARDWARE}

Most of experiments were conducted on a computer that consists of 2 Intel Xeon Gold CPUs, and 4 NVIDIA V100 GPUs. We deploy only one experiment on an entire GPU. For the computation overhead of local training at client, we estimate the relative performance (on average) of CADIS over those one of FedAvg using the same device setting, e.g., the GPU Force-GTX 3090.

Note that some of our experiments are intensive. With aforementioned system, an experiment takes up to \textbf{36 hours} to complete. Therefore, low-resource systems might take few days. We highly recommend a system that is equal to or better than aforementioned settings for experimental reproducibility.

\subsection{SOFTWARE}
We use a pytorch-implemented framework for experimental evaluation. Whilst the full requirement can be viewed in the README.md in Artifact 1, below are some prerequisites:
\begin{itemize}
    \item \textbf{Compilers}: Python 3.8.12
    \item \textbf{Frameworks}: Pytorch - py3.8\_cuda11.3\_cudnn8.2.0\_0
    \item \textbf{Support libraries}: wandb 0.13.5
    \item \textbf{Core libraries}: please check README.md
    \item \textbf{Datasets}: CIAFR-10, MNIST, CIFAR-100.
\end{itemize}

\subsection{ARTIFACTS}

\textbf{Artifact 1}\\
Github: \href{https://github.com/AIoT-Lab-BKAI/ORO-CCGRID2023-CADIS}{ORO-CCGRID2023-CADIS} \\
Artifact name: ORO-CCGRID2023-CADIS\\

\textbf{Artifact 2}\\
Github: \href{https://github.com/WwZzz/easyFL}{easyFL}\\
Artifact name: easyFL\\

% and pill-dataset (which will be conviniently installed when run the experiments. However, the dataset is quite heavy ($\approx 3$ Gb)).

% \section{Theoretical analysis}
% \label{subsec:theory}
% \input{pen_proof1}
% \input{pen_proof2}
% \input{theo-Qmatrix}
% \input{theo-Loss}
\end{document}